\documentclass{article}

\PassOptionsToPackage{numbers,sort&compress}{natbib}
\usepackage[preprint]{neurips_2026}

\usepackage[utf8]{inputenc}
\usepackage[T1]{fontenc}
\usepackage{hyperref}
\usepackage{url}
\usepackage{booktabs}
\usepackage{amsfonts}
\usepackage{amsmath}
\usepackage{nicefrac}
\usepackage{microtype}
\usepackage{xcolor}
\usepackage{graphicx}
\usepackage{multirow}
\usepackage{array}
\usepackage{algorithm}
\usepackage{algpseudocode}

\newcommand{\method}{SAE-Audit}
\newcommand{\flash}{deepseek-v4-flash}
\newcommand{\terra}{GPT-5.6-terra}
\newcommand{\luna}{GPT-5.6-luna}
\newcommand{\ci}[2]{[$#1$, $#2$]}

\title{SAE-Audit: Clinical-Trial Prediction via Inference-Time Optimization of Black-Box LLMs Audited by Sparse Autoencoder Evidence}

\author{%
  Junbo Shen \\
  Independent Researcher \\
  \texttt{jacobjunboshen@gmail.com} \\
}

\begin{document}

\maketitle

\begin{abstract}
Clinical trials are costly and often take years to complete, making early outcome prediction valuable for trial planning. Large language models (LLMs) are increasingly used to forecast clinical-trial outcomes, yet their predictions remain unreliable and their rationales may not faithfully reflect the evidence underlying their answers. On tasks such as predicting whether a registered clinical trial will complete or terminate early, frontier LLMs perform only modestly above chance in our experiments. We introduce \method{}, which optimizes a black-box LLM at inference time with evidence obtained from a deployable open model. A frozen sparse autoencoder (SAE) on Llama-3.1-8B-Instruct is converted into a vocabulary of clinical concepts. Sparse probes fitted on past trials determine which concepts are associated with the outcome. For a new protocol, the concepts detected by the probe are written as evidence, each with the direction and strength of its association with the outcome, and the LLM being audited optimizes its answer in the light of this evidence. Without additional training, \method{} raises balanced accuracy of \flash{} by 9.6--12.0 points on five clinical-trial prediction tasks and raises \terra{} by 8.0 points on trial completion, exceeding other agentic inference-time optimization methods by a large margin. As an auxiliary, SAE probes module within SAE-Audit is itself accurate and auditable, providing natural-language explanations directly derived from verified concept contributions and protocol evidence, without requiring additional LLMs at inference time.  \method{} provides verifiable, concept-level evidence for more accurate and auditable clinical-trial prediction. It can also be embedded into biomedical AI agents such as Biomni to optimize their performance.
\end{abstract}

\section{Introduction}

A clinical trial is designed, registered and funded years before anyone learns whether it will be completed, how long it will run, or whether participants will drop out, and the only information available at that point is the protocol, a long free-text document. Large language models (LLMs) read such documents fluently, yet prompting them for the outcome is unreliable. On the trial-completion task of TrialPanorama~\citep{wang2026trialpanorama}, GPT-4o reaches a balanced accuracy of 59.3 and o3-mini 51.0, where 50 is chance. The two models studied here, \flash{} and \terra{}, reach 61 and 63. Each answer from LLMs comes with a fluent rationale, but such rationales often hallucinate or misrepresent how a model arrived at its answer~\citep{turpin2023language,lanham2023measuring}, and we cannot faithfully check them in clinical trial studies.

\begin{figure}[t]
  \centering
  \includegraphics[width=\linewidth]{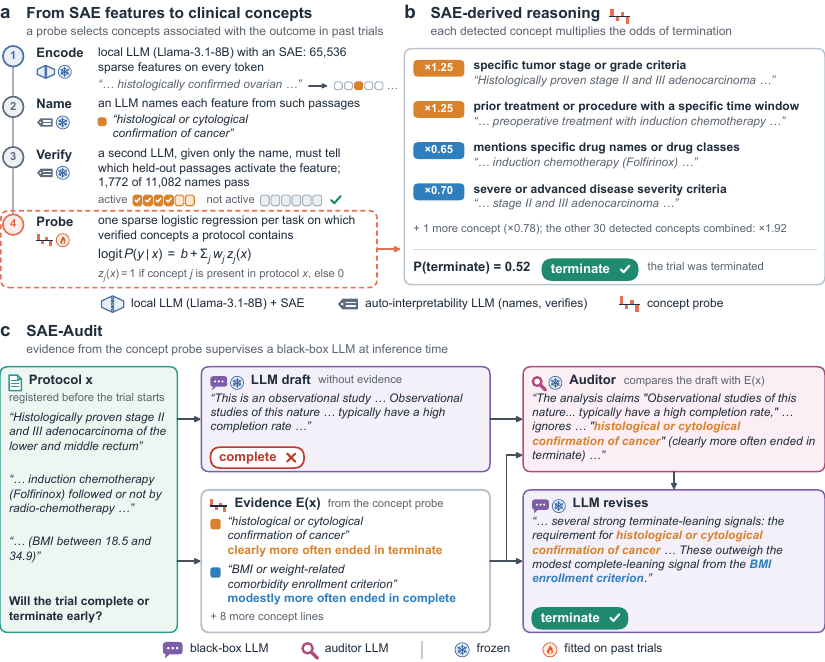}
  \caption{\textbf{Overview of \method{}}, with SAE concept dictionary build-up, SAE probes module which can serve as both a direct reasoner and predictor, and SAE-Audit that takes evidence from SAE probes for inference-time optimization on LLMs. \textbf{a}, A frozen SAE on Llama-3.1-8B-Instruct is turned into a vocabulary of clinical concepts, and one sparse probe per task is fitted on which concepts past protocols contain. \textbf{b}, SAE-derived reasoning: the probe's prediction is an exact product of factors, one per detected concept, each tied to the passage of the protocol on which the concept is most active. \textbf{c}, \method{} in the draft-first setting with the auditor: the concepts detected by the probe are written as evidence $E(x)$; the auditor checks the LLM's draft against the evidence, and the LLM revises its answer, here from \emph{complete} (wrong) to \emph{terminate} (correct).}
  \label{fig:method}
\end{figure}

Two lines of work address parts of this problem. The first improves an LLM's answer at inference time through critique and revision~\citep{madaan2023selfrefine,shinn2023reflexion,du2024improving,yuksekgonul2025textgrad}. When the critique comes from the model itself or from another LLM, it adds no information from outside the models, and self-correction of this kind has been found not to improve reasoning unless external feedback is available~\citep{huang2024large,gou2024critic}. The second reads and uses language models' internal activations. A linear probe on hidden activations predicts completion sometimes better than prompted LLMs (Appendix~\ref{app:probe}), but it outputs a score that cannot be inspected. Sparse autoencoders (SAEs) decompose activations into features that can be described in words~\citep{cunningham2024sparse,gao2025scaling,lieberum2024gemmascope,marks2025sparse,karvonen2025saebench}, and recent methods turn such descriptions into hypotheses for human researchers~\citep{movva2025hypothesaes,simon2025interplm}. However, in prior SAE works, the explanations or descriptions are the end product; they are not used to inference-time refine or optimize LLM or agents.

\method{} connects the two (Fig.~\ref{fig:method}). It converts what a probe on a small open model has learned from past trials into evidence that a black-box LLM can check, and uses this evidence to refine the LLM at inference time. A frozen SAE~\citep{goodfire2024sae} on Llama-3.1-8B-Instruct~\citep{grattafiori2024llama3} is first turned into a vocabulary of clinical concepts: an LLM names each feature, and the name is kept only if a second LLM, given only the name, can identify the text on which the feature is active. One sparse probe per task is then fitted on which of these concepts a protocol contains, so that its prediction is a sum of contributions from named concepts. For a new protocol, the concepts detected by the probe are written as evidence, each with the direction and strength of its association with the outcome in past trials. The evidence then supervises the LLM through its prompt. The LLM either revises its own draft against the evidence or reads the evidence before it answers, and an auditor LLM can first relate the evidence to the protocol. SAE-Audit only trains a simple SAE probe module per task, thus cost-friendly.

On five outcome-prediction tasks from two public clinical-trial benchmarks~\citep{wang2026trialpanorama,chen2025trialbench}, \method{} raises the balanced accuracy of \flash{} by 9.6 to 12.0 points on every task, and that of \terra{} by 8.0 points on trial completion. Two types of control experiments attribute the gain to the content of the evidence. Inverting the direction of every evidence line lowers the accuracy of \flash{} below its original level on every task, and supplying directly the SAE probe's predicted label to LLMs yields only a small or no gain in accuracy. Self-Refine, multi-agent debate and TextGrad, which add no external information, improve accuracy by at most 1.7 points on average, and appending the names of the SAE concepts without their measured associations does not help these methods to improve performance. \method{} also matches the accuracy of Biomni~\citep{huang2025biomni}, a general-purpose biomedical AI agent, with a twentieth of its tokens.

Because the SAE probe is linear in the presence of named concepts, its prediction decomposes exactly into contributions that are each tied to a passage of the protocol. This SAE-derived reasoning requires no further LLMs, quotes the protocol verbatim, and is more faithful under deletion tests than the rationales that the LLMs give for their own answers. When the same probe is fitted on dense neurons or on a random dictionary, the percentage of its features that can be named is roughly halved. The SAE basis is thus what allows the evidence to be put into words. And for prediction performance directly via SAE probe, with continuous activations, the same SAE is on average more accurate on clinical trials classification and regression tasks than other LLMs that we prompt directly. Thus the SAE probes module inside SAE-Audit itself is an accurate and auditable predictor with direct reasoning for its predictions, something other ML or text embedding-based methods cannot provide.

SAE-Audit evidence is generated independently of the target LLM and delivered in natural language. Thus, the framework can also serve as a modular evidence-checking component within existing LLM agents. Such agents can make better and more faithful predictions by incorporating evidence when making or reviewing predictions, without fine-tuning their underlying models.

\section{\method{}}
\label{sec:overview}

\method{} has an offline stage, which builds a vocabulary of verified concepts and one probe per task, and an online stage, which applies them to a new protocol. We describe the method with the example of Fig.~\ref{fig:method}, a study of rectal cancer from the test set of the trial-completion task, and give the details in Appendix~\ref{sec:methods}. Algorithm~\ref{alg:method} summarises both stages.

\begin{algorithm}[t]
  \caption{\method{}}
  \label{alg:method}
  \small
  \algrenewcommand\algorithmicrequire{\textbf{Input:}}
  \algrenewcommand\algorithmicensure{\textbf{Output:}}
  \algrenewcommand{\algorithmiccomment}[1]{\hfill{\color{black!55}$\triangleright$~#1}}
  \algtext*{EndFor}
  \begin{algorithmic}[1]
    \Require a local LLM with a frozen SAE; passages of past protocols; past trials $\{(x_i,y_i)\}$ with known outcomes; a black-box LLM; a new protocol $x$
    \Ensure the answer for $x$
    \Statex \textbf{Verified concepts} (offline, without outcome labels)
    \For{each SAE feature $j$ that is active on enough passages}
      \State $n_j\gets\mathrm{Name}(\text{passages on which $j$ is active},\ \text{passages on which it is not})$ \Comment{an LLM}
      \State $m_j\gets\mathrm{MCC}(\mathrm{Verify}(n_j,\ \text{held-out passages}),\ \text{activations of $j$})$ \Comment{a second LLM, given only $n_j$}
    \EndFor
    \State $C\gets\{j:\ m_j\geq 0.7\}$ \Comment{the concepts}
    \Statex \rule{0pt}{2.9ex}\textbf{Sparse probe over concepts} (offline, once per task)
    \State $z_j(x_i)\gets 1$ if concept $j\in C$ is present in protocol $x_i$, else $0$
    \State $(w,b)\gets$ $L_1$-regularised logistic regression of $y_i$ on $z(x_i)$; \ $S\gets\{j:\ w_j\neq 0\}$ \Comment{Eq.~\ref{eq:probe}}
    \State $\Delta_j\gets\hat P(y{=}1\mid z_j{=}1)-\hat P(y{=}1)$ \ for $j\in S$ \Comment{effect of concept $j$}
    \Statex \rule{0pt}{2.9ex}\textbf{Evidence and audit} (online, for the new protocol $x$)
    \State $E(x)\gets\big[(n_j,\ \operatorname{sign}\Delta_j,\ |\Delta_j|)\big]_{j\in T(x)}\oplus A(x)$ \Comment{Eq.~\ref{eq:evidence}; no LLM call}
    \State $d_0\gets\mathrm{LLM}(x)$ \Comment{draft-first setting only}
    \State $a\gets\mathrm{Auditor}(x,d_0,E(x))$ \Comment{optional}
    \State $d_1\gets\mathrm{LLM}(x,d_0,E(x),a)$ \Comment{Eqs.~\ref{eq:draftfirst} and~\ref{eq:evidencefirst}}
    \State \Return the answer stated in $d_1$
  \end{algorithmic}
\end{algorithm}

\paragraph{Verified concepts.} We pass text through Llama-3.1-8B-Instruct and a frozen SAE self-supervised trained on its layer-19 activations, which yields 65,536 sparse features per token. We give an SAE feature its concept meaning via a LLM auto-interpretability method, using deepseek-v4-flash. An LLM is shown text on which the feature is active and text on which it is not, and proposes a short name such as \emph{histological or cytological confirmation of cancer} (Fig.~\ref{fig:method}a). A name is useful only if it identifies the feature, so each name is tested: a second LLM receives only the name and must pick out, among held-out passages, those on which the feature is active. Of 11,082 named features, 1,772 pass this test with a Matthews correlation of at least 0.7. We call these features \emph{concepts}.

\paragraph{A sparse probe over concepts.} A protocol $x$ is represented by the concepts it contains: $z_j(x)=1$ if concept $j$ is active anywhere in $x$, and $z_j(x)=0$ otherwise. For each task we fit an $L_1$-regularised logistic regression on past trials,
\begin{equation}
  \operatorname{logit} P(y{=}1\mid x)\;=\;b+\sum_{j\in S} w_j\,z_j(x),
  \label{eq:probe}
\end{equation}
where $S$ is the set of concepts with nonzero weight (162 concepts for trial completion). For each concept in $S$ we also measure its \emph{effect} $\Delta_j=\hat P(y{=}1\mid z_j{=}1)-\hat P(y{=}1)$: the outcome rate among training trials that contain the concept, minus the overall rate, estimated on held-out training trials.

\paragraph{Evidence.} The \emph{evidence} $E(x)$ tells the LLM what the probe has learned about the concepts that a new protocol $x$ contains. It is a short text computed from the SAE activations of $x$ and from statistics of the training trials, without any LLM call,
\begin{equation}
  E(x)=\Big[\big(n_j,\ \operatorname{sign}\Delta_j,\ |\Delta_j|\big)\Big]_{j\in T(x)}\ \oplus\ A(x).
  \label{eq:evidence}
\end{equation}
$T(x)$ holds the concepts of $S$ present in $x$ that are most strongly associated with the outcome. Each of them contributes one line that gives its name $n_j$, the outcome it favours, and the size of its effect, in words on a fixed scale or as rates. In Fig.~\ref{fig:method}c, for example, one line reads ``histological or cytological confirmation of cancer: clearly more often ended in terminate''. $A(x)$ states how training trials with a similar set of concepts ended. Every line can be checked: the name of the concept was verified against text, and its effect was measured on past trials.

\paragraph{Audit.} The evidence supervises a black-box LLM at inference time through its prompt; no weights change. Let $\mathrm{LLM}(\cdot)$ denote a call that receives the protocol $x$, the task question and the inputs listed. We use two settings, which differ in when the LLM sees the evidence:
\begin{align}
  \textit{draft-first:}\quad & d_0=\mathrm{LLM}(x),\qquad d_1=\mathrm{LLM}\big(x,\,d_0,\,E(x)\big);\label{eq:draftfirst}\\
  \textit{evidence-first:}\quad & d_1=\mathrm{LLM}\big(x,\,E(x)\big).\label{eq:evidencefirst}
\end{align}
The answer is read from $d_1$. In the draft-first setting, the LLM first answers on its own and then revises its draft $d_0$ against the evidence. Prior work Self-Refine~\citep{madaan2023selfrefine} uses $d_1=\mathrm{LLM}(x,d_0,f)$, with the LLM's critique of its own draft, $f=\mathrm{LLM}(x,d_0)$. We replaced the feedback $f$ with $E(x)$, so the draft is checked against associations measured on past trials rather than against what the LLM already believes. In the evidence-first setting, the LLM reads the evidence before it answers, in a single call. In either setting, an \emph{auditor} can first relate the evidence to the protocol, $a=\mathrm{Auditor}(x,d_0,E(x))$, with no $d_0$ in the evidence-first setting. Its output $a$ is added to the inputs of the final call. The auditor is the same LLM with its own instructions. Each of its remarks must name a concept of $E(x)$, keep the direction and strength stated for it, and quote the text it refers to. The auditor may not predict the outcome. Figure~\ref{fig:method}c shows the draft-first setting with the auditor. The draft argues that observational studies usually complete. The auditor replies that the draft ignores concepts associated with termination, among them the requirement of histologically confirmed cancer, and the LLM changes its answer to \emph{terminate}, which is correct.

\paragraph{SAE-derived reasoning.} Equation~\ref{eq:probe} is linear in binary indicators, so the log-odds of a prediction equal the intercept plus the weights of the concepts that are present. The probe can therefore explain its own prediction without any LLM (Fig.~\ref{fig:method}b). The explanation lists the concepts with the largest weights, each as a factor $e^{w_j}$ on the odds together with the passage of the protocol on which the concept is most active, followed by the combined factor of the remaining concepts.

\section{\method{} improves black-box LLMs}
\label{sec:audit}

We evaluate on five binary outcome-prediction tasks from two public benchmarks: trial completion and sample size from TrialPanorama~\citep{wang2026trialpanorama}, and serious adverse events, trial duration and patient dropout from TrialBench~\citep{chen2025trialbench} (Table~\ref{tab:tasks}, Appendix~\ref{app:data}). The input is the protocol as registered before the trial started. Duration and sample size are numeric, and we predict whether they exceed the training median. Each probe is fitted on 6,000 training trials and tested on 1,000 (1,129 and 995 for sample size). Each LLM experiment uses the same 250 randomly drawn test protocols per task and three runs per condition, and we score the majority answer. We report balanced accuracy ($\times 100$) with 95\% paired bootstrap intervals over protocols.

\begin{table}[t]
  \caption{\textbf{\method{} improves a black-box LLM on clinical trial prediction tasks.} Balanced accuracy ($\times 100$) on 250 random test protocols per task; gains are paired differences between LLM + \method{} and the LLM alone with 95\% bootstrap intervals. Self-Refine, debate, TextGrad, and LLM + SAE concepts (LLM directly given the names of the detected SAE concepts) are inference-time optimization baselines (Fig.~\ref{fig:results}a). Biomni is a general-purpose biomedical AI agent, evaluated on trial completion with a single run per protocol (Appendix~\ref{app:audit}). \emph{Concept Probe} is the SAE probe that supplies the evidence (full test sets in Table~\ref{tab:predict}), and can also give direct predictions via the probe. SAE-Audit is draft-first for trial completion, sample size tasks and evidence-first for the other tasks (Appendix~\ref{sec:methods:audit}).}
  \label{tab:audit}
  \centering
  \footnotesize
  \setlength{\tabcolsep}{2pt}
  \begin{tabular}{@{}lccccccccc@{}}
    \toprule
     & LLM & \multicolumn{3}{c}{Inference-time optimization} & & LLM + SAE & Concept & \textbf{LLM +} & Improvement \\
    \cmidrule(lr){3-5}
    Task & alone & Self-Refine & Debate & TextGrad & Biomni & concepts & probe & \textbf{\method{}} & [95\% CI] \\
    \midrule
    \multicolumn{10}{@{}l}{\emph{LLM: \flash{}}} \\
    Trial completion & 61.0 & 54.7 & 61.4 & 58.2 & 70.9 & 59.4 & 65.2 & \textbf{71.2} & $+10.1$ \ci{+3.6}{+16.9} \\
    Adverse events   & 66.4 & 50.0 & 67.5 & 69.9 & -- & 65.7 & 75.0 & \textbf{77.5} & $+11.0$ \ci{+4.6}{+17.6} \\
    Duration         & 66.0 & 56.8 & 64.5 & 67.2 & -- & 58.9 & 75.4 & \textbf{75.6} & $+9.6$ \ci{+2.9}{+16.2} \\
    Sample size      & 57.2 & 64.7 & 60.8 & 63.5 & -- & 61.1 & 67.5 & \textbf{67.6} & $+10.4$ \ci{+4.0}{+16.2} \\
    Dropout          & 52.8 & 48.9 & 54.2 & 53.2 & -- & 55.0 & 64.2 & \textbf{64.9} & $+12.0$ \ci{+0.7}{+23.6} \\
    \bottomrule
  \end{tabular}
\end{table}

\subsection{SAE-Audit refines LLM on five selected clinical-trial tasks}
\begin{figure}[t]
  \centering
  \includegraphics[width=\linewidth]{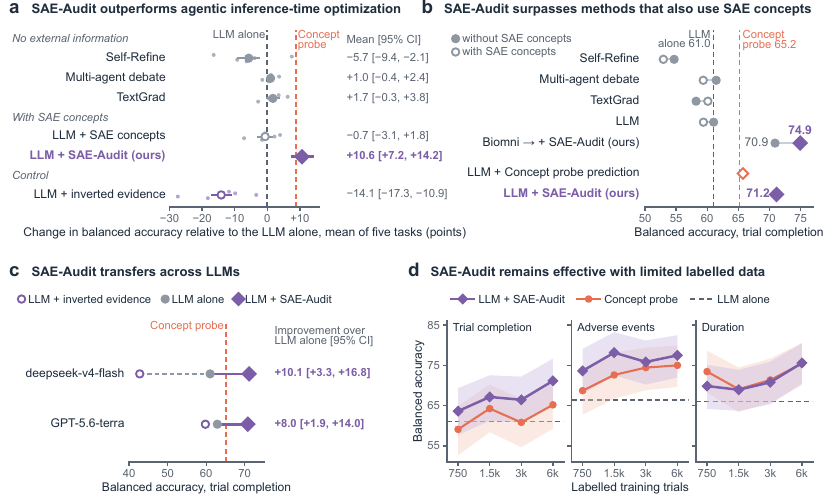}
  \caption{\textbf{\method{} outperforms agentic inference-time optimization methods, surpasses them even when they use SAE concepts, transfers across LLMs and remains effective with limited labelled data.} Balanced accuracy on 250 test protocols per task, with \flash{} unless stated otherwise; intervals and bands are 95\% bootstrap intervals. \textbf{a}, Change relative to the LLM alone, averaged over five tasks (dots: single tasks). \textbf{b}, Trial completion, without and with SAE concepts, and Biomni without and with \method{}. \textbf{c}, Trial completion with two LLMs. \textbf{d}, Concept probe and evidence computed from 750 to 6,000 labelled training trials.}
  \label{fig:results}
\end{figure}
\method{} improves LLM used in our study (\flash{}) on every task (Table~\ref{tab:audit}). Balanced accuracy rises by 9.6 to 12.0 points, and every confidence interval excludes zero. \method{} can improve LLM performance beyond predictions via SAE concept probe, showing that the evidence derived from SAE via SAE-Audit is more important and useful than its own fitted probe. Unaided, the LLM is 8.8 points less accurate than the SAE concept probe on average. Audited, it is more accurate than the SAE concept probe on every task, and clearly more accurate on trial completion (71.2 against 65.2) and adverse events (77.5 against 75.0). Both settings are effective, with and without the auditor. Every combination raises accuracy on every task, and the evidence-first setting without the auditor, a single LLM call, gains 9.1 points on average (95\% CI 5.5--12.8; Appendix~\ref{app:audit}).

\subsection{Improved performance depends on the content of the evidence from SAE-Audit}
\label{sec:controls}
Three controls with the same LLM, prompt and protocols show that the improvement comes from what the evidence says. First, we invert the direction of every evidence line: a concept associated with termination is presented as associated with completion. Accuracy then falls below that of the unaided LLM, by 18.2 points on trial completion, 11.8 on adverse events, 27.5 on duration, 9.9 on sample size and, not significantly, 3.4 on dropout (Fig.~\ref{fig:results}a). The LLM thus follows the evidence, and the gain requires the evidence to be correct. Second, we remove the evidence but keep the instruction to revise. Accuracy is unchanged ($+0.3$ \ci{-0.7}{+1.5} on the full completion test set), so the instruction by itself does not help. Third, we replace the evidence by the probe's predicted label and probability. On trial completion this gives 65.7, no better than the probe itself (65.2) and well below the 71.2 obtained with the evidence (Fig.~\ref{fig:results}b). The named SAE concepts and our SAE-Audit therefore carry information and capability that the probe prediction does not.

Fig.~\ref{fig:results}a and Table~\ref{tab:baselines} compare sources of feedback. Self-Refine~\citep{madaan2023selfrefine}, multi-agent debate~\citep{du2024improving} and TextGrad~\citep{yuksekgonul2025textgrad} run the same kind of step as \method{}, a critique followed by a revision, with feedback that the LLM generates itself. They change accuracy by $-5.7$, $+1.0$ and $+1.7$ points on average and improve the LLM reliably only on sample size. When the feedback is the names and descriptions of the detected SAE concepts, without their measured directions and strengths, the LLM does not improve either ($-0.7$ points, against $+10.6$ with the evidence via SAE-Audit). Neither can Self-Refine, debate, and TextGrad improve the LLM in this setting with additional concept information (Fig.~\ref{fig:results}b). What we need is thus the measured association of each concept with the outcome like SAE-Audit.

We also compare \method{} with Biomni~\citep{huang2025biomni}, a general-purpose biomedical AI agent used as a baseline by TrialAtlas~\citep{lin2026trialatlas}. Biomni achieves a balanced accuracy of 70.9 on trial completion and uses about 146,000 tokens per protocol. LLM + \method{} achieves 71.2 with about 6,900 tokens, a twentieth as many (Table~\ref{tab:audit}, Fig.~\ref{fig:results}b). \method{} also improves Biomni. When Biomni gives different answers to the same protocol, \method{} audits the reasoning behind each answer with its concept-level evidence and selects the best-supported answer. Biomni then achieves 74.9 ($+4.0$ \ci{-0.6}{+8.8}), the highest accuracy on trial completion in our experiments (Fig.~\ref{fig:results}b).

\subsection{SAE-Audit transfers across LLMs}
\label{sec:transfer}

The evidence does not depend on the LLM that receives it. We therefore gave the evidence for the same 250 trial-completion protocols to \terra{}, in the draft-first setting with the auditor (Fig.~\ref{fig:method}c), followed by a selection step (Appendix~\ref{sec:methods:audit}). \method{} raises its balanced accuracy from 62.9 to 70.8 ($+8.0$ \ci{+1.9}{+14.0}; Fig.~\ref{fig:results}c). The stronger model uses the evidence more selectively: when revising, it changes its answer on 18\% of protocols, against 41\% for \flash{}, and two thirds of its changes are corrections. With inverted evidence it gains nothing (63.7 when revising and 59.8 after selection, against 62.9 unaided), whereas \flash{} loses 8.1 points in the same setting (Appendix~\ref{app:audit}).

\subsection{SAE-Audit works under limited-sample settings with fewer labelled trials}
\label{sec:labels}

With fewer labelled trials the probe weakens. We refit the completion probe on 750, 1,500 and 3,000 of the 6,000 training trials and recompute the evidence from the same trials (Fig.~\ref{fig:results}d). Fitted on 750 trials, the probe is less accurate than the unaided LLM (59.1 against 61.0), whereas the audited LLM is more accurate than both (63.6). At each of the four numbers of labelled trials the audited LLM is more accurate than the probe, by 4.7 points on average (95\% CI $-0.1$ to 9.6). The same holds on adverse events (3.6 points on average), whereas on duration the audited LLM and the probe are level ($-1.0$). Over the three tasks the average difference is 2.4 points (95\% CI 0.0 to 4.8; Appendix~\ref{app:audit}). This shows our SAE-Audit can effectively boost LLM’s performance even
with fewer labelled training trials under limited-sample settings.

\section{SAE probes module behind SAE-Audit evidence is faithful, interpretable}
\label{sec:probe}

The SAE-Audit evidence that improves the LLMs in Section~\ref{sec:audit} is produced by SAE probes module. This section tests the module on its own, without any LLM. We ask whether its explanation reflects what drives its own prediction (Section~\ref{sec:explain}), whether its concepts can be named only because the representation is an SAE (Section~\ref{sec:basis}), and how accurate its predictions are (Section~\ref{sec:predict}).

\begin{figure}[t]
  \centering
  \includegraphics[width=\linewidth]{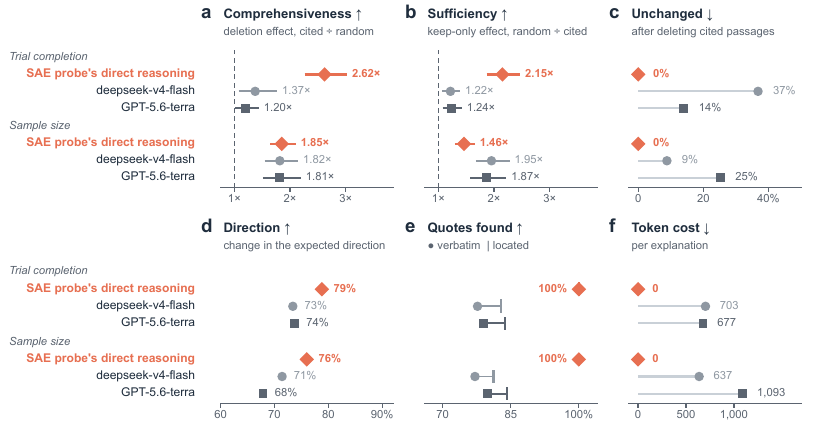}
  \caption{\textbf{SAE probes module's direct reasoning is faithful, fully grounded in the protocol, and cost-friendly} SAE probes module in SAE-Audit can give direct reasoning, comparable or better than LLM explanations, on 250 test protocols per task. \textbf{a}, Change in the prediction after deleting the cited passages, divided by the change after deleting as many random passages, with 95\% bootstrap intervals; $1\times$ is no better than random passages. \textbf{b}, Change after keeping only random passages, divided by the change after keeping only the cited passages. \textbf{c}, Protocols on which deleting the cited passages leaves the prediction unchanged. \textbf{d}, Protocols on which deleting the cited passages changes the prediction in the expected direction. \textbf{e}, Quotations found in the protocol word for word (verbatim) or with small differences in wording (located). \textbf{f}, Token cost: tokens generated per explanation. SAE probes module’s direct reasoning requires no additional LLM.}
  \label{fig:explain}
\end{figure}

\subsection{SAE-derived reasoning is faithful and grounded}
\label{sec:explain}

For each prediction, SAE concept probe directly reports its five largest contributions $w_j z_j(x)$ in Eq.~\ref{eq:probe} and the combined contribution of all other concepts (a complete example is in Appendix~\ref{app:explain}). We compare these explanations with rationales produced by the two LLMs on the same 250 protocols, for trial completion and for the numeric estimation of sample size. Each LLM is prompted for a prediction and up to five pieces of evidence with verbatim quotations. Let $f(x)$ be a method's own prediction for protocol $x$, $r$ the passages it cites, and $\tilde r$ a random set of passages matched in number and length (Appendix~\ref{sec:methods:explain}). Following ERASER~\citep{deyoung2020eraser} and later work on faithfulness~\citep{jacovi2020towards,madsen2024self}, the \emph{comprehensiveness} and \emph{sufficiency} of the cited passages are the mean changes in the prediction when they are deleted and when only they are kept,
\begin{equation}
  \mathrm{comp}(r)=\mathbb{E}_x\,\big|f(x)-f(x\setminus r)\big|,\qquad
  \mathrm{suff}(r)=\mathbb{E}_x\,\big|f(x)-f(r)\big|,
  \label{eq:faith}
\end{equation}
with $\mathrm{comp}(\tilde r)$ and $\mathrm{suff}(\tilde r)$ defined in the same way for the random passages. An explanation is faithful if the prediction depends on the passages it cites. Deleting these passages should then change the prediction more than deleting random passages. Keeping only these passages should change the prediction less than keeping only random passages. Fig.~\ref{fig:explain}a and b therefore show the ratios $\mathrm{comp}(r)/\mathrm{comp}(\tilde r)$ and $\mathrm{suff}(\tilde r)/\mathrm{suff}(r)$, which are above 1 for a faithful explanation. Deleting the cited passages should also change the prediction, $f(x\setminus r)\neq f(x)$. When it does, the change should be in the expected direction. For example, deleting evidence for termination should lower the predicted probability of termination. An explanation is grounded if each quoted passage can be found.

On trial completion the probe's explanation is more faithful than either LLM's rationale (Fig.~\ref{fig:explain}; all values in Table~\ref{tab:explain}). Deleting the passages cited by the probe changes its predicted probability 2.6 times as much as deleting random passages. For \flash{} and \terra{} the ratios are 1.4 and 1.2. For 37\% of protocols, \flash{} returns exactly the same probability after its own cited evidence has been removed. Deleting the passages cited by the probe changes its prediction in the expected direction on 79\% of protocols for trial completion and 76\% for sample size, against 68--74\% for the LLMs. On sample size all three methods have ratios between 1.8 and 1.9, so the probe is not more faithful there. On both tasks the probe's explanations are fully grounded: every passage it cites is verbatim, whereas 16--19\% of the quotations given by the LLMs cannot be located in the protocol. The probe also generates no tokens for its explanation, against 640 to 1,090 for the LLMs (Fig.~\ref{fig:explain}f).

\subsection{Interpretable concepts require the SAE basis}
\label{sec:basis}

Explanation and evidence from SAE-Audit both rely on interpretable features in representations. To verify whether this requires an SAE, we repeat the whole procedure with two other sets of features computed from the same layer-19 activation $h$ of each token: the dense neurons, $\mathrm{ReLU}(h)$, and a \emph{random dictionary} of the same size as the SAE but without training, $\mathrm{ReLU}(Rh)$ with a random matrix $R$~\citep{cunningham2024sparse}. For the probe, we keep only the $k$ largest activations of each token: $k=24$ for the SAE and the random dictionary, and $k=2$ for the 4,096 neurons. Details are in Appendix~\ref{app:basis}.

\begin{figure}[t]
  \centering
  \includegraphics[width=\linewidth]{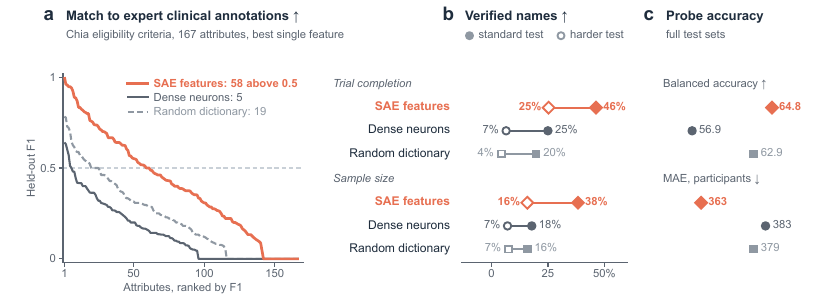}
  \caption{\textbf{SAE features match expert clinical annotations and receive verifiable concept names far more often than dense neurons or random dictionary.} \textbf{a}, Held-out $F_1$ of the best single feature of each basis for each of 167 clinical attributes, such as pregnancy or allergy, that experts annotated in the eligibility criteria of the Chia corpus, ranked. \textbf{b}, Percentage of concept names, written by an LLM, with which a second LLM identifies the passages where the feature is active (MCC $\geq 0.7$), for the features that each probe selects. In the standard test, these are the passages on which the feature is most active. In the harder test, they also include passages on which it is only weakly active. \textbf{c}, Accuracy of probes of different types of features on full test set.}
  \label{fig:basis}
\end{figure}

We first compare the features with clinical concepts annotated by experts. The Chia corpus annotates the eligibility criteria of 1,000 trials with attributes such as pregnancy, allergy or alcohol abuse~\citep{kury2020chia}. For each of 167 attributes, we choose the single feature of each basis that best detects it on half of the trials and score it on the other half (Fig.~\ref{fig:basis}a). SAE features detect 58 attributes with an $F_1$ above 0.5, against 19 for the random dictionary and 5 for dense neurons.

We then fit the same probe on each basis and let the same LLM write a concept name for every feature that the probe selects. As in Section~\ref{sec:overview}, a second LLM sees only the name and must pick out the passages on which the feature is active. In the standard test, these passages are those on which the feature is most active. In a harder test, they also include passages on which the feature is only weakly active. SAE features pass both tests far more often than neurons and random directions (Fig.~\ref{fig:basis}b). On trial completion, 46\% of the SAE features pass the standard test, against 25\% of neurons and 20\% of random directions. In the harder test, 25\% of the SAE features pass, against 7\% of neurons and 4\% of random directions. The results for sample size are similar. SAE features also score much higher in two other standard tests of interpretability~\citep{paulo2024automatically,han2025sage,karvonen2025saebench} (statistical tests in Appendix~\ref{app:basis}). In simulation test, LLM predicts activations of a feature from its concept name. In generation test, a feature should be more active on passages that an LLM writes for its own concept name than on passages written for other names.

Probing via SAE basis does not reduce accuracy (Fig.~\ref{fig:basis}c). On trial completion, the SAE probe is 7.9 points more accurate than dense neuron probe and at least as accurate as the random-dictionary probe. On sample size, its mean error is the lowest of the three.

\begin{table}[t]
  \caption{\textbf{Predictive accuracy on test sets via only the SAE probes module in SAE-Audit.} Balanced accuracy ($\times 100$) for 5 classification tasks; sample size is also scored as mean absolute error (MAE) in participants as regression-style tasks. The concept probe uses binary indicators of verified concepts and supplies the explanations and the SAE-Audit evidence.}
  \label{tab:predict}
  \centering
  \footnotesize
  \setlength{\tabcolsep}{2.6pt}
  \begin{tabular}{@{}lcccccc@{}}
    \toprule
    & Completion & Duration & Adverse events & Dropout & \multicolumn{2}{c}{Sample size (995)} \\
    \cmidrule(lr){6-7}
    Method & (1,000) & (1,000) & (1,000) & (1,000) & $>$ median & regression MAE \\
    \midrule
    \multicolumn{7}{@{}l}{\emph{Prompted directly}} \\
    \flash{}                                            & 61.2 & 63.6 & 64.1 & 54.9 & 60.6 & 422 \\
    \luna{}                                             & 64.5 & 66.5 & \textbf{81.1} & 55.3 & \textbf{77.1} & 409 \\
    \midrule
    \multicolumn{7}{@{}l}{\emph{SAE probes module}} \\
    Concept probe                                       & 64.8 & 74.3 & 75.6 & 64.9 & 64.0 & 363 \\
    Auxiliary prediction probe (fully trained)                         & \textbf{69.3} & \textbf{76.2} & 80.3 & \textbf{67.0} & 74.1 & \textbf{337} \\
    \bottomrule
  \end{tabular}
\end{table}

\subsection{On its own, SAE probes module rivals prompted LLMs and explains its predictions}
\label{sec:predict}

Besides supplying the SAE-Audit evidence, the module can predict trial outcomes and explain its predictions by itself, with one forward pass of an 8B model and no LLM call (Table~\ref{tab:predict}). The 8B model can be hosted locally, so an unregistered draft protocol can be assessed without being sent to an external service. The concept probe uses binary indicators of verified concepts, which make its explanation exact (Section~\ref{sec:explain}) and supply the SAE-Audit evidence. An auxiliary prediction probe on the continuous activations of the named SAE features is more accurate, but its predictions cannot be explained. Averaged over the five tasks, the auxiliary prediction probe reaches a balanced accuracy of 73.4, against 60.9 for \flash{} and 68.9 for \luna{}. \luna{} is more accurate only on adverse events and sample size. The auxiliary prediction probe is as accurate as a probe on the dense activations of the same layer, so the SAE basis costs no accuracy (Appendix~\ref{app:probe}). The concept probe is more accurate than \flash{} on every task. On trial completion, both probes also exceed the published results of GPT-4o (59.3) and four other LLMs~\citep{wang2026trialpanorama} (Appendix~\ref{app:probe}). On the numeric sample-size task, the concept probe and the auxiliary prediction probe have mean errors of 363 and 337 participants, against 409 to 422 for 2 LLMs.

\section{Discussion}
\label{sec:discussion}

\method{} improves an LLM by giving it information that it does not have and that it can check. Without the evidence, the instruction to revise changes nothing, in line with the finding that LLMs do not improve their reasoning by self-correction without external feedback~\citep{huang2024large}. With inverted evidence the LLM loses accuracy (Section~\ref{sec:controls}).

SAE probes module in the SAE-Audit framework itself is an accurate predictor whose reasoning can be inspected without extra LLM calls (Section~\ref{sec:probe}), and the SAE-Audit shows that what the probe has learned can be passed on in words. An LLM that is nine points less accurate than the SAE concept probe exceeds the probe's accuracy by reading the evidence derived from the probe on different tasks.

Methods that describe SAE features in words, such as HypotheSAEs and InterPLM, present the descriptions to human researchers as hypotheses~\citep{movva2025hypothesaes,simon2025interplm}. We instead use SAEs to change and refine the output of another LLM. Verbalisation methods train a model to describe activations in natural language~\citep{pan2026latentqa,karvonen2025activation,li2026verbalization}. In \method{} no language models are trained, and we get verified SAE concepts via auto-interpretability using API models. TextGrad improves a system by propagating an LLM's critique as text~\citep{yuksekgonul2025textgrad}. In our SAE-Audit, the feedback comes from evidence derived from SAE concept probes whose concept associations were measured on past trials.

\section{Limitations}
\label{sec:limitations}

The LLM experiments use 250 random test protocols per task instead of the full test set, due to heavy cost of the LLM API calls, and for the TrialBench tasks the test sets are themselves 1,000-trial samples of the official splits, so the confidence intervals are wide. The setting of the audit shown for each task is the most suitable one on its evaluation protocols. A single setting for all tasks gains 9.1 points on average (Appendix~\ref{app:audit}). Concept names were verified on eligibility-criterion passages only (Fig.~\ref{fig:basis}b). Finally, all results rely on one SAE at one layer of one 8B model, and the probe captures associations in registered trials, not causes of trial outcomes. The clinical utility of \method{} remains to be evaluated prospectively.

\section{Conclusion}

We show that SAE features from a frozen, general-purpose, locally deployable 8B language model can be used for more than post-hoc interpretation. Once named, verified, and connected to historical trial outcomes, they form an auditable predictor and an external source of evidence for black-box frontier LLMs. A sparse concept probe decomposes each prediction into signed contributions traceable to protocol passages, while the associated outcome statistics allow a black-box LLM to improve its forecast without fine-tuning. Across five clinical-trial prediction tasks, this feedback significantly improves LLMs' answering, well beyond inference-time refinement by self-generated critique, debate or TextGrad.

SAE-Audit therefore connects interpretability and LLM inference-time adaptation: the same concept space supports accurate and auditable prediction, inspection of predictive evidence, and targeted supervision of other LLMs. For clinical-trial forecasting and planning, this offers an alternative to task-specific LLM fine-tuning: a frozen open model and a lightweight probe per task provide evidence that an off-the-shelf LLM can use and clinical researchers can verify against the protocol. More broadly, our results demonstrate that SAEs built on deployable language models can go beyond mechanistic interpretability and explaining model predictions: the verified concepts they uncover can serve as actionable, evidence-grounded supervision that improves the predictions of other models, including frontier LLMs, at inference time. Moreover, SAE-Audit is suitable for integration into existing LLM-based clinical-research agents as an external prediction and auditing component.

\bibliographystyle{unsrtnat}
\bibliography{references}

\appendix

\section{Method details}
\label{sec:methods}

\subsection{Tasks and data}
\label{sec:methods:data}

Table~\ref{tab:tasks} (Appendix~\ref{app:data}) lists the tasks and splits. Every method receives the same text for a protocol, cut at 3,000 characters; the probe reads its first 512 tokens and the LLMs all of it. For TrialPanorama, the text consists of the registered fields of the protocol: title, design, eligibility criteria and arm groups for completion, and the benchmark's description of the trial for sample size. TrialBench describes a trial in several modalities: free text, drug molecules, disease codes and tabular features of the design. We use a processed version of the benchmark, in which the molecules appear as SMILES strings next to the MeSH terms of the interventions and the tabular features as lines such as ``Drug intervention Number: 2.0'', and join eight of its fields into one text; the disease codes are not used (Appendix~\ref{app:data}).

\subsection{Concept vocabulary}
\label{sec:methods:dict}

We use the pretrained SAE on layer 19 of Llama-3.1-8B-Instruct, which has 65,536 features~\citep{goodfire2024sae}; neither model is trained. Names are derived from 20,000 eligibility-criterion passages taken from 1,791 training protocols of the completion task, without outcome labels. A feature that is active on enough passages is named and tested as in Section~\ref{sec:overview}, and the answers of the second LLM are scored against the true activations with the Matthews correlation coefficient (MCC)~\citep{matthews1975comparison}. Of 11,082 named features, 1,772 reach an MCC of at least 0.7 and form the vocabulary, to which we add 107 names obtained in the same way from sample-size protocols (Appendix~\ref{app:dict}). Both LLMs are \flash{} at temperature 0. About 15\% of the names are near-duplicates, which we keep.

\subsection{Per-task Probing}
\label{sec:methods:probe}

A concept is present in a protocol, $z_j(x)=1$, if its feature is among the 24 most active on at least one of the first 512 tokens. For each classification task we fit the $L_1$-regularised logistic regression of Eq.~\ref{eq:probe} on the training split, over the verified concepts. The concepts with nonzero weight form the basis $S$, which has 162 concepts for completion, 186 for duration, 222 for adverse events, 155 for dropout, and 40 for sample size (Appendix~\ref{app:probe}). Sample size is also predicted as a number, with a Lasso regression of the logarithm of enrolment on the same indicators (107 concepts). The effect of concept $j$ is $\Delta_j=\hat P(y{=}1\mid z_j{=}1)-\hat P(y{=}1)$, estimated on one half of the training split; the other half is used to rank the concepts for display (Appendix~\ref{sec:methods:audit}). In the evidence it is expressed on a fixed verbal scale: \emph{overwhelmingly}, \emph{clearly} or \emph{modestly} more often for $|\Delta_j|$ of at least 0.25, 0.12 or 0.06, and \emph{only slightly} more often below that.

\subsection{Evidence, audit and controls}
\label{sec:methods:audit}

The evidence $E(x)$ has three parts. The \emph{concept lines}, $T(x)$ in Eq.~\ref{eq:evidence}, are the present concepts among the 24 concepts of $S$ most strongly associated with either outcome on training trials (twelve per outcome, by log-odds ratio), followed by further present concepts in order of their association with the outcome until there are at least ten lines with the auditor and four without. Each line gives the name of the concept, a one-sentence description, the outcome it favours and its effect. The \emph{task summary} is a short description of the probe, written once per task by an LLM from the names of the concepts most strongly associated with each outcome. The \emph{aggregate} states how many of these 24 concepts are present in $x$ for each outcome, and how training trials with a similar set of concepts ended.

In the draft-first setting (Eq.~\ref{eq:draftfirst}), the LLM receives the protocol, the task question, its draft and the feedback, and is asked to revise its analysis in the light of the feedback. In the evidence-first setting (Eq.~\ref{eq:evidencefirst}), it receives the protocol, the feedback and the task question. The feedback is $E(x)$ and, with the auditor, the auditor's output: remarks on the draft or, in the evidence-first setting, a few hypotheses about the trial, each supported by a phrase quoted from the protocol and by a concept of $E(x)$. Appendices~\ref{app:audit} and~\ref{app:prompts} give the wording of $E(x)$ in each setting and the prompts. For \flash{}, the setting of each task (Table~\ref{tab:audit}) is the one with the highest accuracy on the evaluation protocols of the task; Appendix~\ref{app:audit} reports every setting on every task. \terra{} is audited draft-first with the auditor, followed by a selection step that keeps, when the drafts and the revisions of the three runs disagree, the answer whose analyses are more consistent with the concepts they mention. The inverted control negates every $\Delta_j$, the aggregate and the outcomes of the listed trials, the second control removes $E(x)$ from the prompt, and the third replaces $E(x)$ by the probe's predicted label and probability.

\subsection{Evaluation of explanations}
\label{sec:methods:explain}

Passages are lines of the protocol, with long lines split at sentence boundaries, and a quotation given by an LLM is mapped to the passage that contains it (Appendix~\ref{app:explain}). The prediction $f$ in Eq.~\ref{eq:faith} is the probability of termination, or the logarithm of the estimated enrolment. For the probe, $f$ is recomputed from the edited text; for an LLM, the edited text is submitted again. The random set $\tilde r$ is drawn once per protocol; it has as many passages as $r$ and a total length within 25\% of it.

\section{Tasks, data and splits}
\label{app:data}

\begin{table}[h]
  \caption{\textbf{The five prediction tasks.} The input is the protocol as registered before the trial started.}
  \label{tab:tasks}
  \centering
  \footnotesize
  \setlength{\tabcolsep}{3pt}
  \begin{tabular}{@{}lllll@{}}
    \toprule
    Task & Source & Train / test & Original label & Predicted \\
    \midrule
    Trial completion & TrialPanorama & 6,000 / 1,000 & completed or terminated & same \\
    Sample size      & TrialPanorama & 1,129 / 995   & participants enrolled & number; above median (153)? \\
    Adverse events   & TrialBench & 6,000 / 1,000 & any serious adverse event? & same \\
    Duration         & TrialBench & 6,000 / 1,000 & length in years & above median (1.545 y)? \\
    Dropout          & TrialBench & 6,000 / 1,000 & any dropout? & same \\
    \bottomrule
  \end{tabular}
\end{table}

\paragraph{Serialisation.} Every method reads the same text of a protocol, cut at 3,000 characters. For trial completion, the text is built from the TrialPanorama entry of the protocol: the title; the design (allocation, intervention model, observational model, primary purpose, time perspective, masking, phase and number of arms) as \texttt{key=value} pairs; the eligibility criteria with gender, age range and acceptance of healthy volunteers; and the arm groups. For sample size, it is the description of the trial that the benchmark provides: official title, conditions, arm groups, interventions, primary outcomes and, where given, the statistical assumptions. TrialBench describes a trial by free text, drug molecules, disease codes and tabular features. For its three tasks we use a processed version of the benchmark distributed on the Hugging Face Hub (\texttt{araag2/TrialBench}), in which these modalities are stored as text, and join eight of its fields as labelled lines in this order: title, summary, condition, phase, eligibility criteria, interventions (names, descriptions, MeSH terms and the SMILES strings of the drug molecules), study design, and additional information, 26 tabular features such as the number of arms of each type, the number of interventions of each type and the class of the lead sponsor, each written as ``name: value''. The ICD-10 codes of the conditions and the keyword lists of the release are not used. Because the fields keep this order, the interventions are present in 50--57\% of the texts and the additional information in 33--40\%.

\paragraph{Splits.} The TrialPanorama completion split has 42,195 training and 1,000 test trials; we use the full test set and 6,000 randomly sampled training trials. Its sample-size split has 1,143 and 1,000 trials; we use 1,129 and 995 after removing records with zero enrolment (test records NCT04992598, NCT05011838, NCT05237219, NCT05427344 and NCT05866237). The TrialBench data contain 14,368 / 3,548 (adverse events), 34,188 / 8,667 (duration) and 30,561 / 7,741 (dropout) training / test trials, from which we sample 6,000 / 1,000. The positive class makes up 48\% of the test set for completion (terminated), 67\% for adverse events, 83\% for dropout, 50.6\% for duration and 49.3\% for sample size. We report balanced accuracy so that the two imbalanced tasks are scored on the same footing as the others.

\paragraph{Training-set size.} Refitting the completion probe on random subsets of 750, 1,500, 3,000 and 6,000 training trials gives balanced accuracies of 59.7, 59.4, 63.2 and 64.0 on the full test set (the fitting procedure is that of Appendix~\ref{app:basis}; the probe used in the paper scores 64.8). Accuracy has not saturated at 6,000 trials.

\paragraph{Evaluation subsets.} The 250 test protocols of an LLM experiment are the first 250 of a fixed random permutation of the test set. On these subsets the probe scores 65.2 (completion), 75.4 (duration), 75.0 (adverse events), 64.2 (dropout) and 67.5 (sample size), against 64.8, 74.3, 75.6, 64.9 and 64.0 on the full test sets.

\section{Concept vocabulary}
\label{app:dict}

Names were derived from 20,000 eligibility-criterion passages of 1,791 trials, which were divided by trial into two halves. A feature counts as active on a passage if it is among the 24 most active features on at least one token, and it is a candidate if it is active on at least 30 passages and at most 25\% of the passages of each half (median 1.2\%). On the first half, an LLM sees the six longest passages on which the feature is active and six on which it is not, and proposes two names. On the second half, a second LLM receives the two names and twelve passages chosen in the same way, six active and six inactive in random order, and lists for each name the passages that have the named property. It sees no activations and none of the passages used for naming. An active passage is shown as an excerpt of up to 19 words, and an inactive passage, drawn at random, by its first 170 characters. The name with the higher MCC is kept, and it passes at an MCC of at least 0.7, which allows one wrong passage among the twelve or two errors of the same kind. Over the 11,082 named features the median detection $F_1$ is 0.600; 3,196 names reach an $F_1$ of 0.70 and 1,772 an MCC of 0.7; of these, 181 classify all twelve passages correctly, 587 make one error and 1,004 two errors of the same kind. The two names are compared on the twelve passages that also give the score; the test of Appendix~\ref{app:basis}, applied to the concepts of each probe, uses separate sets of protocols for naming, for selection and for scoring. A further LLM pass labels 1,600 of these 1,772 as clinical content and 172 as formatting patterns of the registry.

The 107 additional names were obtained for the sample-size task. The 474 features with the largest weights in a preliminary sample-size probe were named on training protocols of that task, with up to three rounds of naming and testing, and 107 names reached an MCC of 0.7 on 16 active and 16 inactive held-out passages; 66 of these features already had a verified name. The choice of these 474 features used the sample-size training labels. Removing the 107 names changes the MAE of the sample-size probe from 363.4 to 364.5.

\section{Probe details and variants}
\label{app:probe}

\paragraph{Fitting.} For trial completion, duration, adverse events and dropout, the probe is an $L_1$-regularised logistic regression (inverse regularisation strength 0.05) over the 1,600 verified concepts labelled as clinical content. All concepts with nonzero weight are kept, and the decision threshold is the median training score. For sample size, which has 1,129 training trials, the candidates are the verified concepts present in 1--50\% of training protocols. The regularisation is relaxed until at least 40 concepts have nonzero weight; the 40 with the largest weights are kept and refitted with an $L_2$ penalty, and the threshold is the out-of-fold score quantile at one minus the training positive rate. The numeric sample-size probe is a Lasso regression of log enrolment, with the penalty chosen by five-fold cross-validation; it keeps 107 concepts and has an intercept of 5.22, that is, 185 participants.

\paragraph{Auxiliary prediction probe.} The auxiliary prediction probe of Section~\ref{sec:predict} uses three statistics per named feature (maximum, mean, and share of tokens on which the feature is active, over 1,024 tokens), ranks features by training AUROC, and fits a ridge-regularised logistic regression; the number of features and the penalty are chosen jointly by five-fold cross-validation. Table~\ref{tab:ridge} compares it with a probe on the dense layer-19 activations of the same model and with a probe on MedCPT embeddings~\citep{jin2023medcpt}, both fitted in the same way. The auxiliary prediction probe matches the dense probe: all four paired differences have intervals that include zero. Its five largest contributions account for only 2--3\% of a prediction, which is why the explanations and the evidence use binary indicators. On the numeric sample-size task the auxiliary prediction probe reaches an MAE of 336.8; the concept probe reaches 363.4, and predicting the training median 382. \flash{}, asked for the planned sample size (median of three runs), reaches 421.8, and \luna{} (one run) 408.6; an answer without a number was asked once more for a best estimate, which concerns 67 of the 2,985 answers of \flash{} and one answer of \luna{}. On dropout, which is not in Table~\ref{tab:ridge}, the auxiliary prediction probe scores 67.0 against 64.9 for the concept probe.

\begin{table}[h]
  \caption{\textbf{Probe variants and probe baselines} (balanced accuracy $\times 100$ on the full test sets). The difference is between the first two rows, paired over test trials, with a 95\% bootstrap interval. The first row is the auxiliary prediction probe and the last row the concept probe.}
  \label{tab:ridge}
  \centering
  \footnotesize
  \setlength{\tabcolsep}{4pt}
  \begin{tabular}{@{}lcccc@{}}
    \toprule
    & Completion & Duration & Adverse events & Sample size \\
    \midrule
    SAE concepts, continuous & 69.3 & 76.2 & 80.3 & 74.1 \\
    Dense activations & 70.4 & 74.4 & 81.0 & 72.1 \\
    Difference [95\% CI] & $-1.1$ \ci{-3.7}{+1.5} & $+1.9$ \ci{-0.4}{+4.1} & $-0.7$ \ci{-2.6}{+1.3} & $+2.0$ \ci{-0.5}{+4.4} \\
    \midrule
    MedCPT embeddings & 65.0 & 73.0 & 75.8 & 67.4 \\
    SAE concepts, binary & 64.8 & 74.3 & 75.6 & 64.0 \\
    \bottomrule
  \end{tabular}
\end{table}

\paragraph{Published results.} Table~\ref{tab:published} lists the results that TrialPanorama reports for prompted LLMs on the same test sets~\citep{wang2026trialpanorama}.

\begin{table}[h]
  \caption{\textbf{Published results of prompted LLMs} on the same test sets~\citep{wang2026trialpanorama}, next to the two SAE probes of Table~\ref{tab:predict}. Balanced accuracy ($\times 100$) on trial completion and mean absolute error in participants on sample size.}
  \label{tab:published}
  \centering
  \footnotesize
  \begin{tabular}{@{}lcc@{}}
    \toprule
    & Completion (1,000) & Sample size, MAE (995) \\
    \midrule
    GPT-4o                                & 59.3       & 452 \\
    o3-mini                               & 51.0       & --  \\
    GPT-4o-mini, Qwen3-8B, Llama-3-70B    & 51.3--52.6 & --  \\
    \midrule
    Concept probe                         & 64.8       & 363 \\
    Auxiliary prediction probe            & \textbf{69.3} & \textbf{337} \\
    \bottomrule
  \end{tabular}
\end{table}

\section{Explanations}
\label{app:explain}

\paragraph{The explanation of the protocol in Fig.~\ref{fig:method}} (trial completion, test item 10), in full. Verdict: terminate (P(terminate) $=0.52$). Evidence for terminate: ``prior treatment or procedure with a specific time window'' ($\times 1.25$; detected at: ``Patient undergoing preoperative treatment with induction chemotherapy (Folfirinox) followed or not by radio-chemotherapy of the Cap50 type''); ``specific tumor stage or grade criteria'' ($\times 1.25$; detected at: ``Histologically proven stage II and III adenocarcinoma of the lower and middle rectum''). Evidence for complete: ``mentions specific drug names or drug classes'' ($\times 0.65$; detected at: ``Patient undergoing preoperative treatment with induction chemotherapy (Folfirinox) followed or not by radio-chemotherapy of the Cap50 type''); ``severe or advanced disease severity criteria'' ($\times 0.70$; detected at: ``Histologically proven stage II and III adenocarcinoma of the lower and middle rectum''); ``listing specific cancer subtypes or anatomical sites'' ($\times 0.78$; detected at: ``Limited legal capacity or legal incapacity''). The other 30 detected concepts lean, on balance, toward terminate ($\times 1.92$). The 5 concepts above carry 42\% of the evidence. The intercept of this probe is zero, so the odds are the product of the six factors, $1.25\times1.25\times0.65\times0.70\times0.78\times1.92=1.06$, which gives the probability of 0.52.

\paragraph{A sample-size example} (test item 198). Estimate: 99 participants. Evidence for a larger trial: ``Specifies quantitative thresholds or measurement methods'' ($\times 1.15$; detected in an arm description: ``Participants receive non-resuscitation fluids according to local routines, with the following stipulations: Maintenance fluids (crystalloids and/or glucose and/or enteral water) are given at\ldots''). Evidence for a smaller trial: ``feasibility study design with no formal statistical assumptions'' ($\times 0.78$; detected at: ``Official Title: Protocolized Reduction of Non-resuscitation Fluids Versus Usual Care in Septic Shock Patients: A Multicentre Feasibility Trial''); ``trials labelled feasibility or pilot'' ($\times 0.80$; same title); ``specifies a numeric quantity of discrete components'' ($\times 0.85$; detected at: ``\ldots continued for the duration of the ICU admission up to a maximum of 90 days''); ``mentions specific drug names or classes in eligibility criteria'' ($\times 0.91$; detected at: ``\ldots glucose at a concentration of at least 20\% and a maximal dose of 1g/kg/day may be used as nutrition if enteral feeding is not tolerated''). The other 19 detected concepts lean, on balance, toward a smaller trial ($\times 0.97$). The 5 concepts above carry 58\% of the evidence. The estimate is $185\times1.15\times0.78\times0.80\times0.85\times0.91\times0.97=99.6$, where 185 corresponds to the intercept.

\paragraph{Evaluation details.} On average the five listed concepts account for 38\% (completion) and 52\% (sample size) of the total absolute weight of the concepts present in a protocol. A quotation given by an LLM is mapped to a passage by exact match or, failing that, by fuzzy match with a similarity of at least 0.85. Eleven answers of \flash{} on edited sample-size protocols at first contained no number and were obtained by asking once more for a best estimate. One rationale of each LLM contains no quotation that can be located; it is scored as an empty rationale and remains in every statistic. The random sets match the cited ones in number of passages and have 0.80--0.89 of their length. Intervals use 5,000 bootstrap resamples.

\paragraph{Further statistics for Table~\ref{tab:explain}.} On trial completion, deleting the cited passages changes the prediction by 0.133 for the probe, against 0.051 for random passages; the values are 0.152 against 0.110 for \flash{} and 0.109 against 0.091 for \terra{}. Keeping only the cited passages changes it by 0.073 against 0.157 (probe), 0.226 against 0.275 (\flash{}) and 0.140 against 0.174 (\terra{}). Among protocols on which cited and random passages give different results, the cited passages have the larger deletion effect in 77\%, 61\% and 57\% of cases; the two are identical for 0\%, 34\% and 11\% of protocols. On sample size the deletion effects are 0.346 against 0.187 (probe), 0.512 against 0.282 (\flash{}) and 0.442 against 0.244 (\terra{}) in log enrolment, and the keep-only effects 0.247 against 0.360, 0.345 against 0.673 and 0.363 against 0.679. The probe takes 0.4--0.5 s per protocol (one forward pass and one SAE encoding), and the LLMs 8--12 s.

\begin{table}[h]
  \caption{\textbf{Faithfulness and grounding of direct SAE-derived explanations against LLM-generated explanations for the final predictions} (250 test protocols per task): the probe's SAE-derived reasoning against the rationales of two LLMs. Comprehensiveness and sufficiency are the changes in a method's own prediction when the passages it cites are deleted, or are the only passages kept; both are given as the paired difference to a matched random set of passages with 95\% bootstrap intervals. \emph{Unchanged}: the fraction of predictions that remain unchanged after removing cited passages. \emph{Direction}: the percentage of protocols on which deleting the cited passages changes the prediction in the expected direction, among those on which it changes. \emph{Verbatim / located}: the fractions of cited quotations that exactly match or can be found in the original protocol.}
  \label{tab:explain}
  \centering
  \footnotesize
  \setlength{\tabcolsep}{2.3pt}
  \begin{tabular}{@{}lcccccc@{}}
    \toprule
     & $\uparrow$ & $\uparrow$ & $\downarrow$ & $\uparrow$ & \multicolumn{2}{c}{Quotations $\uparrow$} \\
    \cmidrule(lr){6-7}
    Method & Comprehensiveness & Sufficiency & Unchanged & Direction & verbatim & located \\
    \midrule
    \multicolumn{7}{@{}l}{\emph{Trial completion}} \\
    \textbf{SAE probe's direct reasoning} & $\mathbf{+0.082}$ [0.069, 0.096] & $\mathbf{+0.084}$ [0.068, 0.099] & \textbf{0\%} & \textbf{78.8\%} & \textbf{100\%} & \textbf{100\%} \\
    \flash{} & $+0.041$ [0.009, 0.074] & $+0.049$ [0.016, 0.082] & 37\% & 73.4\% & 77.7\% & 82.9\% \\
    \terra{} & $+0.019$ [0.001, 0.037] & $+0.034$ [0.012, 0.054] & 14\% & 73.7\% & 79.1\% & 83.6\% \\
    \midrule
    \multicolumn{7}{@{}l}{\emph{Sample size}} \\
    \textbf{SAE probe's direct reasoning} & $+0.159$ [0.127, 0.192] & $+0.114$ [0.077, 0.151] & \textbf{0\%} & \textbf{76.0\%} & \textbf{100\%} & \textbf{100\%} \\
    \flash{} & $\mathbf{+0.230}$ [0.164, 0.298] & $\mathbf{+0.329}$ [0.254, 0.404] & 9\% & 71.4\% & 77.1\% & 81.2\% \\
    \terra{} & $+0.198$ [0.135, 0.268] & $+0.316$ [0.224, 0.401] & 25\% & 67.9\% & 79.9\% & 84.1\% \\
    \bottomrule
  \end{tabular}
\end{table}

\paragraph{Origin of the cited passages.} On trial completion, 57\% of the passages cited by the probe are eligibility criteria or summary text, 15\% are design fields, 12\% arm descriptions and 16\% titles or headers. The name of a concept shares no content word with its cited passage in 84\% of lines on completion and 65\% on sample size, against 14--24\% for the claims and quotations of the LLMs: a concept is detected from activations, not from matching words. The predictions of the LLMs are coarse. \flash{} uses 14 distinct probability values, all multiples of 5, and 78--82\% of the LLMs' sample-size estimates are multiples of 10, which explains why their predictions are often identical before and after an edit. The probe cites one passage per concept although 42\% (completion) and 34\% (sample size) of the cited concepts are also active elsewhere in the protocol, so the deletion test understates its comprehensiveness.

\paragraph{The largest weights of the probes.} Table~\ref{tab:global} lists the ten concepts with the largest weights in each probe. The twenty largest weights account for 40\% of the total absolute weight on completion and 49\% on sample size.

\begin{table}[h]
  \caption{\textbf{The ten concepts with the largest weights in each probe}, with their factor and the share of test protocols in which they are present.}
  \label{tab:global}
  \centering
  \footnotesize
  \begin{tabular}{@{}p{6.3cm}lcc@{}}
    \toprule
    Concept & Favours & Factor & Present \\
    \midrule
    \multicolumn{4}{@{}l}{\emph{Trial completion} (162 concepts; factor on the odds of termination)} \\
    mentions specific drug names or drug classes & complete & 0.65 & 78\% \\
    specifies a threshold or range for a lab value or test result & complete & 0.68 & 25\% \\
    severe or advanced disease severity criteria & complete & 0.70 & 33\% \\
    time intervals after procedures or events & terminate & 1.34 & 33\% \\
    concurrent medication or trial participation restrictions & complete & 0.75 & 16\% \\
    listing specific cancer subtypes or anatomical sites & complete & 0.78 & 74\% \\
    mentions follow-up or cross-over procedures & complete & 0.79 & 19\% \\
    mentions specific drug names or drug combinations & complete & 0.80 & 35\% \\
    prior treatment or procedure with a specific time window & terminate & 1.25 & 17\% \\
    specific tumor stage or grade criteria & terminate & 1.25 & 10\% \\
    \midrule
    \multicolumn{4}{@{}l}{\emph{Sample size} (107 concepts; factor on the estimated enrolment)} \\
    criteria for classifying an event as an endpoint & larger & 1.29 & 24\% \\
    feasibility study design with no formal statistical assumptions & smaller & 0.78 & 11\% \\
    death mentioned as an event in the outcome measure or in the statistical assumptions & larger & 1.27 & 4\% \\
    pilot study label & smaller & 0.80 & 10\% \\
    trials labelled feasibility or pilot & smaller & 0.80 & 12\% \\
    titles naming trials in low- or middle-income countries with community-based delivery & larger & 1.25 & 9\% \\
    exclusion for neurological or CNS conditions & smaller & 0.81 & 15\% \\
    device interventions involving physical apparatus or technology & smaller & 0.82 & 14\% \\
    describes specific clinical signs or symptoms & smaller & 0.82 & 12\% \\
    intervention protocols with precise numeric parameters & smaller & 0.82 & 27\% \\
    \bottomrule
  \end{tabular}
\end{table}

\section{Comparison of bases}
\label{app:basis}

\paragraph{Features.} For the layer-19 activation $h$ of a token, the SAE features are $\mathrm{ReLU}(W_{\mathrm{enc}}h+b_{\mathrm{enc}})$, the dense neurons are $\mathrm{ReLU}(h)$, and the random dictionary is $\mathrm{ReLU}(Rh)$. The matrix $R$ has the same size as $W_{\mathrm{enc}}$, $65{,}536\times4{,}096$, and its rows are Gaussian random vectors scaled to length one, drawn once. On every token, about half of the random directions have a positive activation, because a random direction points toward $h$ as often as away from it. For all three sets, we keep only the $k$ largest activations of each token, with $k=24$ for the SAE and the random dictionary and $k=2$ for neurons, which keeps about the same fraction of each set. A feature is present in a protocol if it is kept on at least one token. For each basis we fit the probe of Appendix~\ref{sec:methods:probe}.

\paragraph{Naming and testing.} Every selected feature is named and tested on text from three disjoint sets of training protocols. On the first, an LLM sees six passages on which the feature is most active and six on which it is inactive, and proposes two names. On the second, the name that better separates six active from six inactive passages is selected. On the third, a second LLM that sees only the selected name classifies ten active and ten inactive passages, and we record the MCC; the active passages are either those on which the feature is most active (standard test) or are drawn evenly from weak to strong activation (harder test). For simulation, the second LLM rates twenty passages from 0 to 10 given the name, and we report the Pearson correlation with the true activations. For generation, an LLM writes five passages from the name, and we report the AUROC with which the feature's activation separates them from the passages written for all other names of the same probe. Percentages are compared with two-proportion $z$-tests and per-feature scores with one-sided Mann--Whitney tests.

\paragraph{Number of features.} For neurons and for the random dictionary, the $L_1$ penalty is relaxed until at least as many features as in the SAE probe have nonzero weight; the features with the largest weights are kept and refitted, and the threshold is the out-of-fold score quantile at one minus the training positive rate. On trial completion this gives 162 random directions and 92 neurons, since only 96 neurons are present in 1--50\% of training protocols. Fitting the SAE concepts in this way gives a balanced accuracy of 64.0, against 64.8 for the probe used in the paper, which is the one reported in Table~\ref{tab:basis}. On sample size the Lasso keeps 107 SAE concepts, 28 of 73 eligible neurons, and 408 random directions. Table~\ref{tab:basis} reports the 107 random directions with the largest weights, so that the number of features matches that of the SAE; Table~\ref{tab:basisfull} also gives all 408.

\paragraph{Additional measures.} Table~\ref{tab:basisfull} adds a stricter threshold for verification and the accuracy of each probe. We also computed a thresholded generation accuracy, the percentage of generated passages on which a feature is active above half of its maximum on example passages. This measure does not separate the bases, because neurons and random directions are active on almost any text: when each feature is scored on passages written for a different feature's name, the accuracy is still 36--66\% for these two bases, against 2--4\% for the SAE. The AUROC of the main text avoids this problem.

\begin{table}[h]
  \caption{\textbf{Interpretability comparison against different types of embeddings.} The same probe selects $K$ features from each basis, and an LLM writes a concept name for each feature. \emph{Verified}: percentage of concept names with which a second LLM identifies the passages where the feature is active (MCC $\geq 0.7$), tested on the passages on which the feature is most active (\emph{standard} test) or on a set that also includes passages on which it is only weakly active (\emph{harder} test). \emph{Median MCC}: median MCC of the harder test. \emph{Simulation $r$}: correlation between the activations an LLM predicts from the concept name and the true activations. \emph{Generation AUROC}: AUROC of the feature's activation on passages an LLM writes for its own concept name versus for other concept names.}
  \label{tab:basis}
  \centering
  \footnotesize
  \setlength{\tabcolsep}{4pt}
  \begin{tabular}{@{}llcccccc@{}}
    \toprule
     & & & \multicolumn{2}{c}{Verified $\uparrow$} & & & \\
    \cmidrule(lr){4-5}
    Task & Basis & $K$ & standard & harder & Median MCC $\uparrow$ & Simulation $r$ $\uparrow$ & Generation AUROC $\uparrow$ \\
    \midrule
    \multirow{3}{*}{Completion} & \textbf{SAE features}   & 162 & \textbf{46\%} & \textbf{25\%} & \textbf{0.44} & \textbf{0.55} & \textbf{0.76} \\
                                & Dense neurons           & 92  & 25\% & 7\%           & 0.31          & 0.40          & 0.68          \\
                                & Random dictionary       & 162 & 20\% & 4\%           & 0.23          & 0.43          & 0.71          \\
    \midrule
    \multirow{3}{*}{Sample size} & \textbf{SAE features}  & 107 & \textbf{38\%} & \textbf{16\%} & \textbf{0.50} & \textbf{0.52} & \textbf{0.74} \\
                                 & Dense neurons          & 28  & 18\% & 7\%           & 0.35          & 0.35          & 0.56          \\
                                 & Random dictionary      & 107 & 16\% & 7\%           & 0.30          & 0.36          & 0.68          \\
    \bottomrule
  \end{tabular}
\end{table}

\begin{table}[h]
  \caption{\textbf{Comparison of bases, additional columns.} \emph{Pool}: features present in 1--50\% of training protocols. \emph{Verified}: percentage of selected features whose concept name reaches an MCC of at least 0.7 in the standard test / at least 0.9 in the standard test / at least 0.7 in the harder test. \emph{Median MCC}: standard test / harder test. \emph{Accuracy}: balanced accuracy on the full test set (completion) or MAE (sample size).}
  \label{tab:basisfull}
  \centering
  \scriptsize
  \setlength{\tabcolsep}{3.5pt}
  \begin{tabular}{@{}llccccccc@{}}
    \toprule
    Task & Basis & Pool & $K$ & Verified & Median MCC & Simulation $r$ & Generation & Accuracy \\
    \midrule
    Completion & SAE features & 1,520 & 162 & 46\% / 23\% / 25\% & 0.65 / 0.44 & 0.55 & 0.76 & 64.8 \\
    Completion & Dense neurons & 96 & 92 & 25\% / 8\% / 7\% & 0.47 / 0.31 & 0.40 & 0.68 & 56.9 \\
    Completion & Random dictionary & 38,115 & 162 & 20\% / 6\% / 4\% & 0.47 / 0.23 & 0.43 & 0.71 & 62.9 \\
    \midrule
    Sample size & SAE features & 1,462 & 107 & 38\% / 15\% / 16\% & 0.58 / 0.50 & 0.52 & 0.74 & 363 \\
    Sample size & Dense neurons & 73 & 28 & 18\% / 7\% / 7\% & 0.47 / 0.35 & 0.35 & 0.56 & 383 \\
    Sample size & Random dictionary & 35,958 & 107 & 16\% / 4\% / 7\% & 0.42 / 0.30 & 0.36 & 0.68 & 379 \\
    Sample size & Random dictionary, all & 35,958 & 408 & 14\% / 3\% / 5\% & 0.35 / 0.30 & 0.32 & 0.67 & 361 \\
    \bottomrule
  \end{tabular}
\end{table}

\paragraph{Statistical tests.} We compare the SAE with each alternative. For the percentage of verified names in the standard test, the two-proportion $z$ statistic on trial completion is 5.1 against the random dictionary ($p\approx4\times10^{-7}$) and 3.4 against neurons ($p\approx8\times10^{-4}$); on sample size it is 3.7 against the random dictionary ($p\approx10^{-4}$). For the percentage in the harder test, $z$ is 5.3 and 3.7 on completion; on sample size it is 1.9 against the random dictionary ($p=0.03$) and 1.2 against neurons ($p=0.12$, with only 28 neurons). One-sided Mann--Whitney tests on per-feature scores give, on completion, $p=1.7\times10^{-10}$ and $5.6\times10^{-6}$ for the MCC across the range (random dictionary, neurons), $2.6\times10^{-4}$ and $2.8\times10^{-4}$ for simulation, and $1.1\times10^{-2}$ and $1.9\times10^{-3}$ for generation; on sample size, $8.2\times10^{-5}$ and $1.3\times10^{-2}$ for the MCC, $4.6\times10^{-5}$ and $3.1\times10^{-3}$ for simulation, and $1.8\times10^{-2}$ and $6.0\times10^{-5}$ for generation.

\paragraph{Accuracy.} On trial completion the SAE probe is 1.8 points more accurate than the random-dictionary probe \ci{-1.2}{+4.9} and 7.9 points more accurate than the neuron probe \ci{+4.2}{+11.5} (paired bootstrap over test protocols). On sample size its MAE is lower by 15.7 participants than that of the 107 random directions \ci{2.6}{28.8} and by 19.3 than that of the neurons \ci{13.6}{25.0}, and it does not differ from that of all 408 random directions ($+2.1$ \ci{-4.9}{+8.8}).

\paragraph{Human-annotated concepts.} The Chia corpus annotates 1,000 trials with 167 attributes of eligibility criteria~\citep{kury2020chia}. Each basis encodes the same text, the inclusion and exclusion criteria of a trial (up to 1,024 tokens), and a feature's value is its maximum activation over the tokens. For each attribute we choose the single best feature of a basis on half of the trials and score it on the other half. SAE features reach a mean $F_1$ of 0.399, with 58 attributes above 0.5 and 28 above 0.7. Random directions reach 0.221, with 19 and 4 attributes, and neurons 0.137, with 5 and none. The paired difference in $F_1$ is 0.18 \ci{0.14}{0.22} against random directions and 0.26 \ci{0.22}{0.30} against neurons (Wilcoxon $p<10^{-15}$). Restricted to the 11,082 named features of the dictionary, the SAE detects 53 attributes above 0.5.

\section{Audit}
\label{app:audit}

\paragraph{The evidence for the protocol of Fig.~\ref{fig:method}.} The ten concept lines, without their one-sentence descriptions, are listed below; an asterisk marks the concepts among the 24 most strongly associated with either outcome. Names are reproduced as given to the LLM, including truncations.
{\footnotesize
\begin{itemize}\setlength{\itemsep}{0pt}
\item[$\ast$] ``The protocol uses a participant-specific event as the time origin for follow-up.'': modestly more often ended in terminate
\item[$\ast$] ``The protocol requires a washout period free of radiotherapy or chemotherapy prior to enrollment.'': clearly more often ended in terminate
\item[$\ast$] ``The trial is a comparative efficacy or safety study involving at least two proprietary medicinal...'': clearly more often ended in complete
\item ``Prospectively documented stage/grade per classification requ'': clearly more often ended in terminate
\item ``histological or cytological confirmation of cancer'': clearly more often ended in terminate
\item ``The protocol evaluates multiple fixed dose levels of a single investigational product in parallel...'': modestly more often ended in complete
\item ``The protocol mandates exclusion of participants with fixed, non-reversible structural...'': modestly more often ended in terminate
\item ``BMI or weight-related comorbidity enrollment criterion'': modestly more often ended in complete
\item ``Digitally literate with verified device access'': only slightly more often ended in terminate
\item ``The protocol includes a systemic anticancer agent and radiotherapy in concurrent or sequential...'': clearly more often ended in terminate
\end{itemize}}
\noindent The aggregate reads: ``Of the task's twenty-four strongest markers (12 terminate-side, 12 complete-side), this protocol shows 2 terminate-side and 1 complete-side. Recorded registered protocols with a profile like this one clearly more often ended in terminate. The record's lean on this profile is strong and consistent. The most similar recorded cases overwhelmingly more often ended in terminate. The neighborhood's lean is strong.''

\paragraph{Settings.} Table~\ref{tab:forms} reports the two settings of the audit (Appendix~\ref{sec:methods:audit}), each with and without the auditor, for \flash{} on all five tasks. Draft-first with the auditor, the auditor checks the draft against $E(x)$, the LLM rewrites its analysis, and a separate call reads the answer from it (Fig.~\ref{fig:method}c). Draft-first without the auditor, the LLM revises its analysis with $E(x)$ in front of it. This call is the refinement call of the Self-Refine baseline, with $E(x)$ as the feedback. Evidence-first with the auditor, the auditor writes hypotheses that pair a concept with a quoted phrase of the protocol, only where the two match, and the LLM answers with $E(x)$ and these hypotheses. A second instruction for the auditor asks for two to four hypotheses in every case. Evidence-first without the auditor, the LLM receives the protocol and $E(x)$ in a single call. With the auditor, $E(x)$ consists of the concept lines, the task summary and the aggregate of Appendix~\ref{sec:methods:audit}. Without the auditor, the concept lines are the present concepts among the 24 (at least four lines), and $E(x)$ also lists the four most similar training trials with their outcomes; in the draft-first setting, effects and aggregate are then given as percentages and the task summary is omitted. Every combination improves the LLM on every task, by 6.7 to 9.1 points on average, and with inverted evidence none improves on the unaided LLM. Table~\ref{tab:audit} reports each task in its most accurate setting, marked $\dagger$ here; \terra{} is audited draft-first with the auditor and the selection step described below.

\begin{table}[h]
  \caption{\textbf{Both settings of the audit, with and without the auditor, improve the LLM on every task.} Change in balanced accuracy (points) relative to the LLM alone, for \flash{} on the 250 test protocols of each task. Bold: the 95\% interval excludes zero. The first row is the setting of Fig.~\ref{fig:method}c. $\dagger$: the setting used in Table~\ref{tab:audit}. $\ast$: the auditor instruction that asks for two to four hypotheses in every case. Mean: over the five tasks (four in the row marked $\ast$). A dash marks a combination that was not run.}
  \label{tab:forms}
  \centering
  \footnotesize
  \setlength{\tabcolsep}{2.7pt}
  \begin{tabular}{@{}llcccccc@{}}
    \toprule
    Setting & Auditor & Completion & Adverse events & Duration & Sample size & Dropout & Mean [95\% CI] \\
    \midrule
    \multicolumn{8}{@{}l}{\emph{Evidence}} \\
    draft-first & audits the draft & $\mathbf{+8.2}$ & $\mathbf{+7.6}$ & $+5.4$ & $+6.4$ & $+7.2$ & $\mathbf{+7.0}$ \ci{+3.7}{+10.2} \\
    draft-first & none & $\mathbf{+10.1}^{\dagger}$ & $+3.6$ & $\mathbf{+9.0}$ & $\mathbf{+10.4}^{\dagger}$ & $\mathbf{+9.7}$ & $\mathbf{+8.6}$ \ci{+5.5}{+11.7} \\
    evidence-first & writes hypotheses & $+6.2$ & $\mathbf{+8.9}$ & $\mathbf{+9.6}^{\dagger}$ & $\mathbf{+9.0}$ & $\mathbf{+11.4}$ & $\mathbf{+9.0}$ \ci{+5.6}{+12.4} \\
    evidence-first & writes hypotheses$^{\ast}$ & $+4.3$ & $\mathbf{+8.1}$ & $+4.8$ & $\mathbf{+9.6}$ & -- & $\mathbf{+6.7}$ \ci{+3.1}{+10.2} \\
    evidence-first & none & $+7.4$ & $\mathbf{+11.0}^{\dagger}$ & $\mathbf{+8.4}$ & $+6.6$ & $\mathbf{+12.0}^{\dagger}$ & $\mathbf{+9.1}$ \ci{+5.5}{+12.8} \\
    \midrule
    \multicolumn{8}{@{}l}{\emph{Inverted evidence}} \\
    draft-first & audits the draft & $\mathbf{-8.1}$ & $-0.2$ & $\mathbf{-34.8}$ & $\mathbf{-17.3}$ & $-4.8$ & \\
    draft-first & none & $\mathbf{-18.2}^{\dagger}$ & $\mathbf{-31.7}$ & $\mathbf{-32.9}$ & $\mathbf{-9.9}^{\dagger}$ & -- & \\
    evidence-first & writes hypotheses & -- & -- & $\mathbf{-27.5}^{\dagger}$ & -- & -- & \\
    evidence-first & writes hypotheses$^{\ast}$ & $\mathbf{-14.5}$ & $\mathbf{-13.1}$ & $\mathbf{-31.4}$ & $\mathbf{-13.0}$ & -- & \\
    evidence-first & none & $\mathbf{-20.4}$ & $\mathbf{-11.8}^{\dagger}$ & $\mathbf{-30.5}$ & $\mathbf{-15.8}$ & $-3.4^{\dagger}$ & \\
    \bottomrule
  \end{tabular}
\end{table}

\paragraph{Selection step.} In the selection step, the frozen SAE encodes each analysis written by the LLM. An analysis $d$ with answer $\hat y_d\in\{-1,+1\}$ receives the score $\hat y_d\,\langle\omega, c(x)\odot c(d)\rangle/\lVert\omega\odot c(x)\odot c(d)\rVert_1$, where $c(x)$ and $c(d)$ indicate which verified concepts are present in the protocol and in the analysis, and $\omega_j$ is the log-odds ratio of concept $j$ on training trials. The draft and revised analyses of the three runs are pooled; if their answers are not unanimous, the answer with the higher mean score is kept. The step has no fitted parameter.

\paragraph{Draft-first with the auditor (Fig.~\ref{fig:method}c).} Without the selection step, \flash{} improves on all five tasks (Table~\ref{tab:forms}), by 7.0 points on average (95\% CI 3.7--10.2). On trial completion it reaches 69.2 ($+8.2$ \ci{+0.7}{+16.0}); Fig.~\ref{fig:method} shows a protocol from this run. It changes its answer on 103 of 250 protocols, of which 61 are corrections, and with inverted evidence it loses 8.1 points \ci{-14.2}{-1.9}. \terra{} reaches 68.1 ($+5.3$ \ci{-0.0}{+10.7}) and changes its answer on 45 protocols, of which 29 are corrections; with the selection step it reaches 70.8 and changes 61 answers, of which 40 are corrections. With inverted evidence, \terra{} scores 59.8 after selection ($-3.2$ \ci{-9.3}{+3.2}) and 63.7 when only revising, with 50 corrections and 51 new errors.

\paragraph{Dropout.} The dropout probe is fitted as for trial completion and has 155 concepts; it scores 64.2 on the 250 evaluation protocols and 64.9 on the full test set. In the setting of Table~\ref{tab:audit}, evidence-first without the auditor, it changes its answer on 126 protocols, of which 91 are corrections, and gains 12.0 points \ci{+0.7}{+23.6}; with inverted evidence it changes 92 answers, of which 15 are corrections, and scores 3.4 points below the unaided LLM \ci{-11.3}{+4.7}. Draft-first with the auditor, it gains 7.2 points \ci{-1.3}{+15.7}, and 9.0 points with the selection step \ci{-0.3}{+18.2}. With evidence from a probe with 40 concepts, the draft-first setting with the auditor and the selection step scores 63.0 ($+10.2$ \ci{+1.1}{+19.5}).

\paragraph{Full test sets.} Two versions of the draft-first setting without the auditor were run on full test sets of 1,000 trials. One raises balanced accuracy by 6.3 points on trial completion \ci{+3.2}{+9.4}, 6.6 on adverse events \ci{+4.2}{+9.1} and 7.1 on duration \ci{+4.1}{+10.0}, and with inverted evidence accuracy falls by 11.9, 14.5 and 31.8 points. In the other, removing the evidence while keeping the instruction to revise changes accuracy by 0.3 points on trial completion \ci{-0.7}{+1.5}, 0.5 on adverse events \ci{-0.2}{+1.4} and 0.4 on duration \ci{+0.1}{+0.8}.

\begin{figure}[h]
  \centering
  \includegraphics[width=\linewidth]{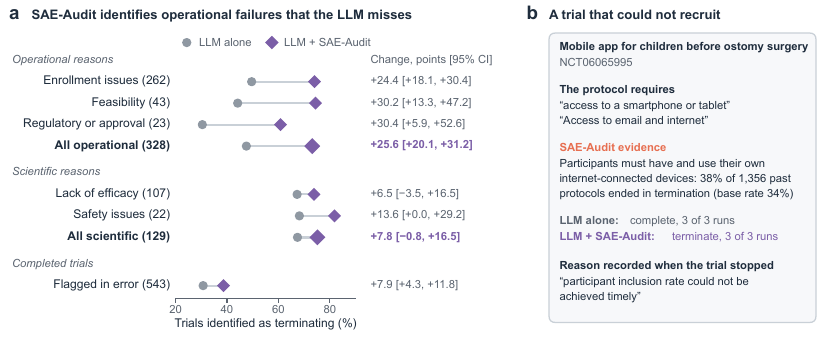}
  \caption{\textbf{\method{} identifies the trials that stop for operational reasons.} \textbf{a}, Percentage of trials identified as terminating by \flash{} on the 1,000 trial-completion test protocols, by the reason recorded for termination. Numbers of trials are in parentheses; intervals are 95\% bootstrap intervals. \textbf{b}, One of the 250 evaluation protocols.}
  \label{fig:reasons}
\end{figure}

\paragraph{Reasons for termination.} TrialPanorama labels each terminated trial with one of five reasons. We call enrollment issues, feasibility and regulatory or approval issues operational, and lack of efficacy and safety issues scientific. They cover 74.8\% and 25.2\% of the 14,804 terminated training trials. Fig.~\ref{fig:reasons}a uses the later version above, with the majority answer of three runs. \flash{} identifies 47.6\% of the trials stopped for operational reasons and 67.4\% of those stopped for scientific reasons. With \method{} it identifies 73.2\% and 75.2\%, changes of $+25.6$ \ci{+20.1}{+31.2} and $+7.8$ \ci{-0.8}{+16.5}. The two changes differ by 17.9 points \ci{+7.5}{+27.6}. On the 750 test protocols outside the 250 evaluation protocols, the changes are $+24.9$ \ci{+18.3}{+31.8} and $+8.3$ \ci{-1.1}{+18.1}. With the setting of Table~\ref{tab:audit} on the 250 evaluation protocols, they are $+28.7$ \ci{+16.7}{+41.5} and $+9.1$ \ci{-10.7}{+29.6}. On completed trials, false alarms rise from 30.8\% to 38.7\%. Among trials stopped for enrollment issues in the 250 protocols, LLM + \method{} identifies 73.4\%, against 56.2\% for Biomni ($+17.2$ \ci{+3.1}{+31.2}).

\paragraph{Fewer labelled trials.} For $N$ labelled trials we take the first $N$ of a fixed random order of the 6,000 training trials of a task. The probe is refitted on these trials with the rule of Appendix~\ref{app:probe} and an inverse regularisation strength of $300/N$, which is 0.05 at 6,000, and every quantity in the evidence (effects, aggregate, task summary and the most similar trials) is computed from the same $N$ trials. The concept vocabulary does not use outcome labels and is unchanged, as are the LLM's drafts and the setting of each task. Table~\ref{tab:labels} gives the results for the three tasks on which the experiment was run. On the full test set of the completion task the probes fitted on 750, 1,500, 3,000 and 6,000 trials keep 279, 283, 236 and 162 concepts and score 61.2, 62.1, 62.5 and 64.9, against 61.2 for the unaided LLM. Averaged over the four values of $N$, with protocols resampled jointly, the audited LLM differs from the probe by $+4.7$ points \ci{-0.1}{+9.6} on trial completion, $+3.6$ \ci{-0.2}{+7.3} on adverse events and $-1.0$ \ci{-4.3}{+2.3} on duration, and by $+2.4$ \ci{+0.0}{+4.8} on average over the three tasks. Its gain over the unaided LLM is 7.1 points on average \ci{+3.7}{+10.4}.

\begin{table}[h]
  \caption{\textbf{Fewer labelled trials.} Balanced accuracy ($\times 100$) on the 250 evaluation protocols when the probe and the evidence are computed from $N$ labelled training trials. The unaided LLM scores 61.0, 66.4 and 66.0 on the three tasks. Differences are paired, with 95\% bootstrap intervals; the rows with 6,000 trials are those of Table~\ref{tab:audit}.}
  \label{tab:labels}
  \centering
  \footnotesize
  \setlength{\tabcolsep}{4pt}
  \begin{tabular}{@{}lrcccc@{}}
    \toprule
    Task & $N$ & Concept probe & LLM + \method{} & Difference to probe [95\% CI] & Gain over LLM [95\% CI] \\
    \midrule
    Trial completion & 750   & 59.1 & 63.6 & $+4.5$ \ci{-2.7}{+12.1} & $+2.6$ \ci{-3.7}{+9.1} \\
                     & 1,500 & 64.3 & 67.1 & $+2.9$ \ci{-4.3}{+9.2}  & $+6.1$ \ci{+0.0}{+12.5} \\
                     & 3,000 & 60.8 & 66.4 & $+5.6$ \ci{-0.9}{+11.9} & $+5.4$ \ci{-0.8}{+11.9} \\
                     & 6,000 & 65.2 & 71.2 & $+6.0$ \ci{-0.5}{+12.3} & $+10.1$ \ci{+3.6}{+16.9} \\
    \midrule
    Adverse events   & 750   & 68.7 & 73.6 & $+4.9$ \ci{-0.6}{+10.7} & $+7.2$ \ci{+0.2}{+14.0} \\
                     & 1,500 & 72.6 & 78.1 & $+5.5$ \ci{+0.1}{+10.8} & $+11.7$ \ci{+5.2}{+18.4} \\
                     & 3,000 & 74.4 & 75.8 & $+1.4$ \ci{-4.0}{+6.8}  & $+9.4$ \ci{+2.8}{+16.0} \\
                     & 6,000 & 75.0 & 77.5 & $+2.5$ \ci{-2.2}{+7.1}  & $+11.0$ \ci{+4.6}{+17.6} \\
    \midrule
    Duration         & 750   & 73.4 & 69.9 & $-3.6$ \ci{-9.6}{+2.1}  & $+3.8$ \ci{-3.2}{+11.0} \\
                     & 1,500 & 69.1 & 68.9 & $-0.1$ \ci{-5.6}{+5.3}  & $+2.9$ \ci{-3.7}{+9.6} \\
                     & 3,000 & 71.4 & 70.8 & $-0.6$ \ci{-5.5}{+4.5}  & $+4.8$ \ci{-2.0}{+11.6} \\
                     & 6,000 & 75.4 & 75.6 & $+0.2$ \ci{-4.3}{+4.5}  & $+9.6$ \ci{+2.9}{+16.2} \\
    \bottomrule
  \end{tabular}
\end{table}

\paragraph{Inference-time baselines.} Table~\ref{tab:baselines} compares \method{} with three methods that revise an LLM's answer at inference time, run with \flash{} on the same 250 protocols per task and starting from the same unaided answers. In \emph{Self-Refine}~\citep{madaan2023selfrefine} the LLM writes feedback on its analysis and then rewrites the analysis; we use one round, which takes the same two calls as an audit. Its refinement call is the final call of the draft-first setting without the auditor: the system prompt, the protocol, the task question, the draft, the instruction and the output limit are the same, and the feedback written by the LLM stands where the audit places $E(x)$. With $E(x)$ the call scores 71.2, 70.1, 75.0, 67.6 and 62.5 on the five tasks, against 54.7, 50.0, 56.8, 64.7 and 48.9 with the LLM's own feedback. The difference is 14.2 points on average \ci{+10.8}{+17.9}, and its interval includes zero only on sample size ($+2.8$ \ci{-4.2}{+9.7}). In \emph{multi-agent debate}~\citep{du2024improving} the three unaided answers act as three agents, and each agent sees the other two answers and gives an updated answer, as in the authors' code (three agents, two rounds). \emph{TextGrad}~\citep{yuksekgonul2025textgrad} is run with the authors' library and its test-time loss: three updates of the first unaided answer and a majority vote over the initial and the three updated solutions. In the second block the names and descriptions of the concepts of $E(x)$ are appended to the protocol, without directions, strengths, aggregate, task summary or similar trials. The LLM gives three new answers on this input, and the three methods start from these answers; they were run on this input for trial completion only. Scores are majority answers over three runs, except for TextGrad, which votes over its own solutions. The baselines are distinguishable from the unaided LLM in their favour only on sample size (Self-Refine $+7.5$ \ci{+1.7}{+13.3}, debate $+3.6$ \ci{+0.5}{+6.8}, TextGrad $+6.3$ \ci{+1.1}{+11.2}), where \method{} remains more accurate. Self-Refine changes the answer on 70 to 197 of the 250 protocols and lowers accuracy on four tasks. On trial completion the first unaided answer scores 59.5 and TextGrad's solution after three updates 55.2. Prompts are in Appendix~\ref{app:prompts}.
\begin{table}[h]
  \caption{\textbf{Inference-time refinement baselines, without and with SAE information, do not close the gap to \method{}.} Balanced accuracy ($\times 100$) of \flash{} on the 250 evaluation protocols of each task, inference-time refinement baselines showing almost no improvements compared to direct LLM answers; the difference to the direct LLM answers averaged over the five tasks with a 95\% bootstrap interval. Fig.~\ref{fig:results}a,b plots these values. After being given the concept names of the protocol via our SAE, baselines still fail to improve the LLM. \emph{Concept names}: the names and descriptions of the SAE concepts detected in the protocol, without their directions and strengths. }
  \label{tab:baselines}
  \centering
  \footnotesize
  \setlength{\tabcolsep}{3.5pt}
  \begin{tabular}{@{}lcccccc@{}}
    \toprule
    Method & Completion & Adverse events & Duration & Sample size & Dropout & Mean diff.\ [95\% CI] \\
    \midrule
    LLM alone & 61.0 & 66.4 & 66.0 & 57.2 & 52.8 & \\
    \midrule
    \multicolumn{7}{@{}l}{\emph{No external information}} \\
    Self-Refine        & 54.7 & 50.0 & 56.8 & 64.7 & 48.9 & $-5.7$ \ci{-9.4}{-2.1} \\
    Multi-agent debate & 61.4 & 67.5 & 64.5 & 60.8 & 54.2 & $+1.0$ \ci{-0.4}{+2.4} \\
    TextGrad           & 58.2 & 69.9 & 67.2 & 63.5 & 53.2 & $+1.7$ \ci{-0.3}{+3.8} \\
    \midrule
    \multicolumn{7}{@{}l}{\emph{With SAE information: concept names appended to the protocol}} \\
    LLM                & 59.4 & 65.7 & 58.9 & 61.1 & 55.0 & $-0.7$ \ci{-3.1}{+1.8} \\
    Self-Refine        & 52.9 & -- & -- & -- & -- & \\
    Multi-agent debate & 59.4 & -- & -- & -- & -- & \\
    TextGrad           & 60.1 & -- & -- & -- & -- & \\
    \midrule
    \multicolumn{7}{@{}l}{\emph{With SAE information: evidence $E(x)$ (ours)}} \\
    LLM + inverted evidence & 42.8 & 54.7 & 38.6 & 47.3 & 49.4 & $-14.1$ \ci{-17.3}{-10.9} \\
    LLM + \method{}    & \textbf{71.2} & \textbf{77.5} & \textbf{75.6} & \textbf{67.6} & \textbf{64.9} & $+10.6$ \ci{+7.2}{+14.2} \\
    \bottomrule
  \end{tabular}
\end{table}

\paragraph{Biomni.} We evaluate Biomni~\citep{huang2025biomni} (version 0.0.8) with \flash{} on the 250 trial-completion protocols, with a single run per protocol and the same system prompt, protocol and task question as the LLM alone. To prevent Biomni from looking up trial outcomes, we remove its literature and database tools (187 of its 235 tools remain), restrict network access to the LLM API and disable shell commands. Each run executes in a sandbox without access to the test labels and is limited to 50 LLM calls or 20 minutes (25 minutes in the second and third runs). Biomni gives no answer on 13 protocols, and for these we use the answer of the LLM alone, as for all other methods. Biomni improves on the LLM alone by 9.9 points \ci{+3.2}{+16.4}. A single run of LLM + \method{} achieves 70.9 on average, and the majority answer of three runs differs from Biomni by $+0.3$ points \ci{-7.1}{+7.8}. To audit Biomni in Section~\ref{sec:controls}, we run it on every protocol twice more with the same setup and without any evidence, and the selection step chooses among the analyses of the three runs. The step has no fitted parameter. Token counts of Biomni are those reported by the API, averaged over its four further runs. For LLM + \method{} they are counted with the DeepSeek tokenizer on the stored prompts and answers of its three runs.

\paragraph{Cost.} An audited protocol requires up to four calls: the draft, which is shared with the unaided condition, the auditor call, the final call of the LLM and, where the answer is read from a revised analysis, one call to read it. The evidence-first setting without the auditor requires one call. Each is repeated for three runs. The task summary requires one call per task. Computing the evidence requires one forward pass of Llama-3.1-8B and no LLM call.

\section{Prompts}
\label{app:prompts}

All LLM calls use temperature 0. Prompts are reproduced verbatim; bracketed terms are filled in per task.

\paragraph{Task summary (once per task).} System: ``You write concise clinical task hypotheses grounded strictly in the marker lists shown. No numbers, no percentages, no invented markers.'' User: ``Registered clinical-trial protocols are indexed by concepts. Trials that ended in [outcome 1] typically show these markers: [eight concepts, each with its name and description]. Trials that ended in [outcome 0] typically show these: [eight concepts]. In THREE sentences, state the task hypothesis: what kind of trial tends to end in [outcome 1], what kind in [outcome 0], and what a reader should look for in a new protocol. Ground every claim in the markers above.''

\paragraph{Auditor, draft-first setting.} System: ``You audit a clinical-trial analysis against a frozen concept detector's reading of the same protocol. You are impartial: you report where the analysis and the record AGREE as well as where they conflict, and the force of your language always matches the recorded strength of the evidence.'' User: the protocol, the draft and the evidence, followed by ``Write an AUDIT of the analysis, 3 to 5 lines, formatted `A1: \ldots'. Each line must pair a specific claim or omission in the analysis (quote its words) with specific concept evidence above (name the concept and carry its recorded direction and strength wording faithfully). Include what the analysis gets RIGHT by the record as well as what it conflicts with or ignores. CALIBRATION RULE: reserve words like `conflicts' or `contradicts' for evidence the record marks `clearly' or stronger; where the record reads `only slightly' or `modestly', present the point as a consideration, not a contradiction. If the record's overall reading is weak or mixed, state that plainly in your first line. Do not predict the outcome; do not tell the analyst what to conclude.''

\paragraph{Auditor, evidence-first setting.} System: ``You are a clinical-trial audit writer. A frozen sparse-autoencoder concept detector, trained only to index registered protocols, has read this protocol. You write instance hypotheses a trialist could check.'' User, for duration: the protocol and the evidence, followed by ``Write INSTANCE HYPOTHESES (`H1: \ldots'), but ONLY where a detected concept and a quoted protocol phrase GENUINELY meet -- the concept's recorded meaning must actually describe that phrase. Never force a pairing; never write a hypothesis to balance the other side. If fewer than two genuine bindings exist, or the genuine ones conflict, write no further hypotheses and say so. Always close with one line `RECORD READING: \ldots' restating the aggregate reading above at exactly its recorded strength wording (both readings separately if they disagree, never blended). Do not predict the outcome; no probabilities.'' User, for sample size: the protocol and the evidence, followed by ``Write 2 to 4 INSTANCE HYPOTHESES about this trial, one line each, formatted `H1: \ldots'. Each must be grounded in BOTH a short quoted phrase from the protocol AND the concepts above (name them). Then add ONE final line `Hs: \ldots' that synthesizes what the detector's aggregate reading says about this protocol, in plain clinical language, keeping its direction and strength faithfully. Do not predict the outcome, do not give probabilities, do not go beyond the listed evidence.''

\paragraph{Final call of the LLM.} Evidence-first without the auditor, the LLM receives the protocol, the evidence and the task question. Draft-first without the auditor, it receives the protocol, the task question, its previous analysis and the evidence, followed by ``Incorporate the feedback above and produce a revised analysis.'', which is also the instruction of the refinement call of Self-Refine below. Draft-first with the auditor, it receives its task prompt, its previous analysis, the evidence and the audit, followed by ``Rewrite your analysis to address the audit -- correct what the record contradicts, keep what your own reading of the protocol genuinely supports, and integrate the named concepts where they sharpen the reasoning. Output ONLY the improved analysis. Do NOT state a final verdict, answer, or outcome.''; the answer is then read from the revised analysis by a separate call that receives the protocol and the revised analysis. Evidence-first with the auditor, it receives the protocol, the evidence, the hypotheses and the task question.

\paragraph{Baselines.} Self-Refine, feedback call. System: ``You are a clinical trial expert reviewing an analysis written for the question below. You give specific, actionable feedback. You do not answer the question yourself.'' User: the protocol, the task question and the analysis, followed by ``Give feedback on this analysis: point out errors in its reasoning, claims that the protocol does not support, and relevant features of the protocol that it overlooks. Be specific and concise. If the analysis needs no change, say so.'' Refinement call: the task prompt, the previous analysis and the feedback, followed by ``Incorporate the feedback above and produce a revised analysis.'' Multi-agent debate: each agent keeps its own answer in the conversation and receives ``These are the solutions to the problem from other agents: [the two other answers] Using the reasoning from other agents as additional advice, can you give an updated answer? Examine your solution and that other agents step by step.'', followed by the answer format of the task. TextGrad (library version 0.1.8): the loss instruction is ``Below is a multi-choice question and an answer. You are an expert scientist. Your job is to investigate the answer. Critically go through reasoning steps, consider your knowledge, and see if the answer is correct or if there are any critical mistakes.'', the variable has the role description ``creative and precise solution and the prediction for the multiple choice question'', the optimiser is constrained to end with the answer format of the task, and all other prompts are the library's. The concept list reads ``A sparse autoencoder on a small language model detects the following concepts in this protocol:'', followed by one line per concept with its name and description.

\paragraph{Label-only control.} The evidence is replaced by ``A frozen predictive model trained on 6,000 registered protocols with known outcomes assigns this trial a [p]\% probability of ending in [outcome 1]; its prediction is [label].''

\paragraph{LLM rationales (Section~\ref{sec:explain}).} Trial completion: ``You are a clinical trial operations expert. Given a trial's protocol as registered BEFORE it ran, judge whether it went on to COMPLETE or was TERMINATED early. Give one or two sentences of reasoning. Then list the evidence you relied on: up to five items, each with (a) a one-sentence factual claim about THIS protocol, (b) a short phrase copied verbatim from the protocol text that supports the claim, and (c) the direction it points to. End with strict JSON on its own line: \{"evidence":[\{"claim":"...","quote":"...","direction":"terminate|complete"\}],"p\_terminate":<integer 0-100>,"outcome":"complete|terminate"\}''. Sample size: the same structure, asking for ``the trial's planned sample size (total number of participants) as its statisticians would have calculated it. For calibration: the median planned sample size of registered trials is 153 participants'', with directions ``larger|smaller'' and an integer ``sample\_size''. When an answer contains no number, the prompt is sent once more with the added line ``A number is required: even if the text gives little basis for a calculation, give your best single estimate in the required JSON.''

\paragraph{Naming and testing of features (Appendix~\ref{sec:methods:dict} and Appendix~\ref{app:basis}).} Naming: ``You are naming one feature of [a sparse autoencoder over clinical-trial protocols / a language model over clinical-trial protocols]. You are shown the text spans where it fires most strongly, and spans where it does not fire. Give TWO different candidate descriptions of the property the firing spans have and the others lack. Each must be ONE short phrase, as SPECIFIC as you can make it while still true of every firing span. A phrase like `requires a specific condition' is useless because it is true of nearly every trial: name the actual requirement, restriction, population, measurement, design element, or manner of writing. Output strict JSON on one line.'' Detection: ``You are given [n] PROPERTIES and numbered clinical-trial text spans. For each property, list the spans that have it, using only the property as stated. Output strict JSON on one line.'' Simulation: ``You are given one PROPERTY and numbered clinical-trial text spans. Rate each span from 0 to 10 for how strongly it exhibits the property, using only the property as stated (0 = not at all, 10 = unmistakably).'' Generation: ``You write text in the style of clinical-trial registry protocols: titles, design lines, eligibility criteria, arm descriptions, outcome measures. Write 5 different short spans (one sentence or one criterion each), each clearly exhibiting the given property. Vary the wording and the clinical context.''

\end{document}